%% file: acm_main.tex
\documentclass[manuscript,screen,10pt]{acmart}
\usepackage{blindtext}
\usepackage{cleveref}
\usepackage[inline]{enumitem}
\usepackage{lscape}
\usepackage{longtable}
\usepackage{makecell}
\usepackage{adjustbox}

\AtBeginDocument{%
  \providecommand\BibTeX{{%
    \normalfont B\kern-0.5em{\scshape i\kern-0.25em b}\kern-0.8em\TeX}}}

\setcopyright{acmcopyright}
\copyrightyear{2022}
\acmYear{2022}
\acmDOI{XXXXXXX.XXXXXXX}

\acmConference[XXX]{Make sure to enter the correct
  conference title from your rights confirmation email}{June 03--05,
  2018}{Woodstock, NY}
\acmPrice{15.00}
\acmISBN{978-1-4503-XXXX-X/18/06}

\begin{document}
\title[A Comprehensive Review of Automated Data Annotation Techniques in HAR]{A Comprehensive Review of Automated Data Annotation Techniques in Human Activity Recognition}

\author{Florenc Demrozi}
\affiliation{%
  \institution{Department of Electrical Engineering and Computer Science, University  of Stavanger}
  \streetaddress{Kitty Kiellands hus, Rennebergstien 30}
  \city{Stavanger}
  \country{Norway}}
\email{florenc.demrozi@uis.no}

\author{Cristian Turetta}
\affiliation{%
  \institution{Department  of Computer Science, University  of Verona}
  \streetaddress{Strada Le Grazie, 15}
  \city{Verona}
  \country{Italy}}
\email{cristian.turetta@univr.it}

\author{Fadi Al Machot}
\affiliation{%
  \institution{Faculty of Science and Technology, Norwegian University  of Life Sciences}
  \streetaddress{Campus Ås, Universitetstunet 3}
  \city{Ås}
  \country{Norway}}
\email{fadi.al.machot@nmbu.no}

\author{Graziano Pravadelli}
\affiliation{%
  \institution{Department  of Computer Science, University  of Verona}
  \streetaddress{Strada Le Grazie, 15}
  \city{Verona}
  \country{Italy}}
\email{graziano.pravadelli@univr.it}

\author{Philipp H. Kindt}
\affiliation{%
  \institution{Faculty of Computer Science, TU Chemnitz}
  \streetaddress{Str. der Nationen 62, 09111}
  \city{Chemnitz}
  \country{Germany}}
\email{philipp.kindt@informatik.tu-chemnitz.de}

\renewcommand{\shortauthors}{Demrozi and Kindt, et al.}
\thanks{\textbf{This is the author’s draft submitted to IEEE/ACM. Copyright may be transferred without notice, after which this version may no longer be accessible. A copyright notice will be added here upon submission, and additional information in the case of an acceptance/publication.}}
\input{sec/0-abstract}

\received{20 December 2022}
\received[revised]{-- ----- 2023}
\received[accepted]{-- ----- 2023}

\maketitle
\input{sec/1-intro}
\input{sec/2-back}
\input{sec/3-search}

\input{sec/4-meth}
\input{sec/5-disc}
\input{sec/6-conc}
\bibliographystyle{ACM-Reference-Format}

\end{document}

%% file: sec/0-abstract.tex
\begin{abstract}
Human Activity Recognition (HAR) has become one of the leading research topics of the last decade. As sensing technologies   have matured and their economic costs have declined, a host of novel applications, e.g., in healthcare, industry, sports, and daily life activities have become popular. The design of HAR systems requires different time-consuming processing steps, such as data collection, annotation, and model training and optimization. 
In particular, data annotation represents the most labor-intensive and cumbersome step in HAR, since it requires extensive and detailed manual work from human annotators. Therefore, different methodologies concerning the automation of the annotation procedure in HAR have been proposed. 
The annotation problem occurs in different notions and scenarios, which all require individual solutions. In this paper, we provide the first systematic review on data annotation techniques for HAR. By grouping existing approaches into classes and providing a taxonomy, our goal is to support the decision on which techniques can be beneficially used in a given scenario.
\end{abstract}



%% file: sec/1-intro.tex
\section{Introduction}\label{sec:intro}
In the last decade, we have witnessed the spread and adoption of sensors, wearables, the Internet of Things (IoT), the Internet of Medical Things (IoMT), and edge computing technologies \cite{li2015internet}. 
Sensors can detect and measure physical properties such as temperature, pressure, light, and motion. They are becoming ubiquitous in various industries, including automotive, aerospace, and consumer electronics. Moreover, their miniaturization has led to their integration into wearables, such as fitness trackers, smartwatches, clothes, and dedicated devices. 
Wearables are frequently used to track various aspects of a person's health and activity. Recent developments even involve integrating medical sensors for remote patient monitoring, digital therapeutics, and real-time intervention into wearables  \cite{seneviratne2017survey,dunn2018wearables,cheng2021recent}. 

On the other side, the IoT is formed by networks of interconnected devices, vehicles, and buildings that communicate with each other and exchange data. It has been adopted across various industries, including home automation, agriculture, and manufacturing. In addition, IoT devices can be remotely monitored and controlled, improving efficiency and productivity \cite{li2015internet}. 
Instead, IoMT refers to using IoT devices in medical applications, enabling the healthcare providers' capacity to monitor patients remotely, collect data for analysis, improve patient outcomes, and reduce healthcare costs \cite{vishnu2020internet}.
Finally, edge computing refers to processing data at or near the source rather than sending it to a central/remote server for processing. This technology has become increasingly important as the amount of data generated by IoT and IoMT devices grows. Edge computing enables faster processing times and reduces the latency and amount of data that needs to be transmitted over the network \cite{baker2023artificial}. 
The adoption and spread of these technologies have revolutionized various industries and enabled new applications and capabilities. With such systems now being ubiquitous, they serve as a common infrastructure for recognizing human activity, as described next.\\

\noindent \textbf{Human Activity Recognition (HAR):} In such a context, HAR is a central research field that finds applications in various areas, including healthcare, sports, industry, and smart homes. HAR refers to the ability to identify and classify human activities using sensors, wearables, or other devices that capture data about the person's movements and actions. With regard to healthcare, HAR can be used to monitor a patients' status and detect abnormalities or changes in their behavior that may indicate a deterioration of health or the onset of a medical condition. For example, HAR can be used to detect falls of elderly patients or to monitor the movements of patients with Parkinson's disease or other motor disorders \cite{demrozi_survey}. 
Moreover, HAR also has applications in sports and fitness to monitor the athletes' performance and technique, helping them to improve their training and prevent injuries. 
HAR can also be used in activity tracking devices, such as fitness trackers, to provide users with insights into their daily activity levels and help them to achieve their fitness goals. 
In addition, HAR automates various tasks in smart homes based on the occupant's activities. For example, lights can be turned on or off automatically based on the person's movements, or the thermostat can be adjusted based on the person's activity level \cite{demrozi_survey,demrozi2021towards}.

HAR is related to various technologies, including sensors, wearables, IoT, IoMT, edge computing, machine learning (ML), Deep Learning (DL), and Artificial Intelligence (AI). Sensors and wearables are used to capture data about the person's movements and actions, which is then used to identify and classify human activities in HAR applications. IoT and IoMT systems are used to collect data from sensors and wearables, which can be transmitted over the network for processing and analysis. Edge computing can process this data at or near the source, reducing latency and enabling real-time processing of HAR data \cite{baker2023artificial,vishnu2020internet}. 

In HAR systems, the data collected from such devices is analyzed to classify a user's activity. While, in principle, this analysis can be done based on heuristics (e.g., a feature exceeds certain thresholds, etc.), ML- and DL-based HAR techniques have become the most popular solution. Using them, also more complex analyses can be carried out, allowing for reliable recognition of activities even in data in which the properties or patterns that represent a certain activity or behavior are not obvious. ML- and Dl-based HAR methods can also integrate other data sources, such as environmental data, to provide more comprehensive insights into human behavior and activity \cite{demrozi_survey}. 
As the technology continues to improve and becomes widely available, we expect to see further advancements and new applications for ML-based HAR \cite{baker2023artificial}. 

\begin{figure*}[!th]
\centering
\includegraphics[width=0.9\textwidth,page={1}]{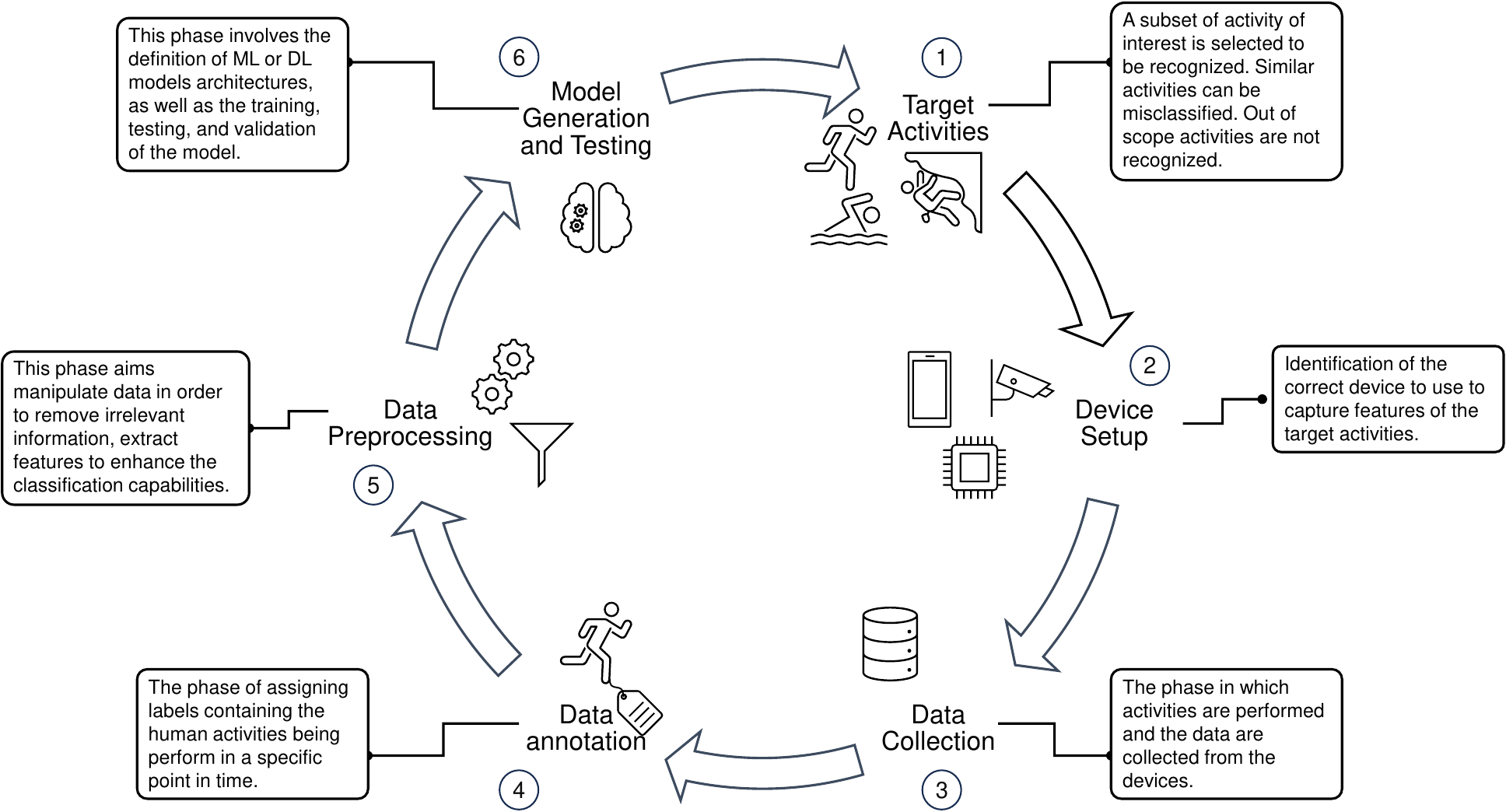}
\caption{Overview on HAR in daily life activities.}  \label{fig:har_overview}
\end{figure*}

When generating HAR model, a set of sensor data is recorded first. This data is then labeled with the activities under consideration. This step is called \textit{annotation}. Next, a machine-learning model is trained, which can then be used to classify unlabeled data. In the following, we describe the individual steps~\cite{demrozi_survey,gupta2022human}
 that are involved in creating a HAR system in more detail. An overview is shown in Figure~\ref{fig:har_overview}.

\begin{itemize}
\item Definition of Target Activities: Definition and analyzation of the real-world characteristics of the target activities to be recognized. For example, this can be their duration, distribution, similarity with other activities, etc. 
\item Device Setup: Identification and study of requirements and determination of the devices to be used in the data collection phase, based on the target human activities.
\item Data Collection: In this phase, data is collected from sensors, wearables, or other devices that capture information about the person's movements and actions.
\item Data Annotation: The process of assigning labels to the human activities being performed. Labels are crucial in supervised learning as they provide the ground truth or correct answers that guide the learning process. By associating input data with corresponding labels, the model can learn to make accurate predictions and generalize its knowledge to unseen examples.
\item Data Preprocessing: The collected data is then preprocessed to remove noise, irrelevant information are filtered out, and the data is prepared for analysis. As a part of this, the following analysis is carried out:
\begin{itemize}
    \item Feature extraction: The preprocessed data is analyzed to extract relevant features that can be used to classify human activities. These features may include movement patterns, body position, or other characteristics.
    \item Feature selection: Once the features have been extracted, a subset of features may be selected for use in the classification model. This helps to reduce the dimensionality (e.g., the number of features) of the data and improve the accuracy of the model.
\end{itemize}
\item Model generation and testing: A HAR (i.e., ML or DL) model is developed to classify human activities based on the selected features in this phase. The model may be trained using a labeled dataset or unsupervised learning techniques. After the model has been generated, the following steps are carried out before the model is ready to be used:
\begin{itemize}
    \item Model evaluation: The developed model is then evaluated using a test dataset to assess its accuracy and performance. This phase helps to identify any issues or areas for improvement in the model.
    \item Deployment: Finally, the developed model is deployed to a real-world environment, where it is used to classify human activities.
\end{itemize}
\end{itemize}

\noindent \textbf{Data Annotation in HAR:} The most labor-intensive step in creating a HAR system is data annotation, which involves creating a labeled dataset for training the ML/DL models. 
Manual labeling, in which human annotators manually label each recorded sample with the corresponding activity, is a common approach in data annotation. Although time-consuming and resource-intensive, it can produce high-quality labels that are accurate and consistent. 
Nevertheless, several factors can pose challenges in the manual data annotation process for HAR systems. \textit{Firstly}, subjectivity can lead to inconsistencies and errors in labeling as the interpretation of the activity being performed can vary among annotators. This can ultimately affect the accuracy of the ML/DL model. 
\textit{Secondly}, the data annotation process can be time-consuming, particularly when labeling large amounts of data, which can cause delays in the development of the HAR system and increase project costs. \textit{Thirdly}, the economic cost can be a limiting factor since hiring human annotators or utilizing crowdsourcing platforms for data labeling can become expensive, mainly when the studied activities are complex. 
\textit{Fourthly}, the variability of human activities can also pose a challenge in the annotation process. Since different individuals can perform activities differently, creating accurate and consistent labels for the data can be challenging. \textit{Lastly}, label noise may exist in annotated data, resulting in errors in the labeling process. Label noise can occur due to human error, subjectivity, or inconsistencies in the annotation process, which ultimately reduces the performance of the HAR system's ML/DL model.
Careful consideration of these limitations and appropriate methods can help mitigate these challenges and improve the accuracy and performance of the final HAR system.

Alternatively, automated methods, such as rule-based systems or unsupervised learning algorithms, can be employed for data annotation. These approaches are more efficient and scalable but may be less precise or necessitate additional manual validation. 
The quality of the annotated data is pivotal to the efficacy of the HAR system. Inaccurate or inconsistent labeling can cause poor ML/DL model performance, leading to the misclassification of human activities \cite{diete2017smart, adaimi2019leveraging}.

There are several (partial) possible solutions to the limitations of the annotation process in HAR \cite{diete2017smart,adaimi2019leveraging}. 
Some of these solutions include:
\begin{itemize}
\item Standardization: Standardizing the annotation process can help to reduce subjectivity and increase consistency in the labeling process. This can be achieved by defining clear guidelines and procedures for annotators to follow and providing training and feedback to ensure the quality of the annotations.
\item Automation: Automated methods, such as unsupervised learning algorithms or rule-based systems, can be used to annotate data. These methods can be faster and more scalable than manual labeling and reduce the annotation process's cost.
\item Active learning: Active learning techniques can reduce the labeled data needed for training an ML or DL model. This involves selecting the most informative data samples for annotation, which can help reduce the labeling process's time and cost.
\item Crowdsourcing: Crowdsourcing platforms can be used to engage many annotators to label the data. This can be a cost-effective solution, as well as provide a diverse range of perspectives on the activity being performed.
\item Quality control: Quality control measures can be implemented to ensure the accuracy and consistency of the labeled data. This can include using multiple annotators to label the same data samples and comparing their annotations, as well as conducting regular checks on the quality of the annotations.
\end{itemize}
While these solutions can enhance the accuracy and performance of the final HAR system, they do not completely eliminate the cost and time needed for the annotation process. \\


\noindent \textbf{Systematic Review Objectives:} This paper aims to systematically review existing methodologies for automating data annotation in HAR. The objective is to identify the strengths and limitations of different techniques and provide insights into the current research and ongoing trends in this area. Specifically, the paper explores different approaches and algorithms used in automatic data annotation techniques. This does not only help in developing novel techniques in the future, but also supports the choice of an appropriate labeling technique for a given application.

This review considers 2401 publications on automating data annotation in HAR. To the best of our knowledge, no systematic review has been published prior to this paper. The  absence of such a review aggravates overseeing the different technologies used in this area, makes it difficult to follow recent trends, and leaves unclear which technical solution is most beneficial for realizing a given scenario.
We in this paper close this gap by providing the first systematic review on this field of research. 
\\

\noindent \textbf{Paper organization:} The rest of the paper is organized as follows. 
Section \ref{sec:back} delves into the background of HAR, presenting a comprehensive overview of the field, including its applications and challenges. 
Following that, Section \ref{sec:search} discusses the selection criteria for annotation methods in HAR, examining the key factors that we consider when choosing appropriate techniques.
Section \ref{sec:meth} presents an in-depth analysis and discussion of various annotation methods employed in HAR, exploring their strengths, limitations, and effectiveness in accurately identifying and classifying human activities 
Finally, Sections \ref{sec:disc} and \ref{sec:conc} conclude the paper by summarizing the key findings and contributions of the study, emphasizing the significance of automatic annotation methods in advancing HAR research and suggesting potential avenues for future exploration in this area.


%% file: sec/2-back.tex
\section{Background}\label{sec:back}
\begin{figure}[!ht]
    \centering
    \includegraphics[width=0.75\textwidth,page={2}]{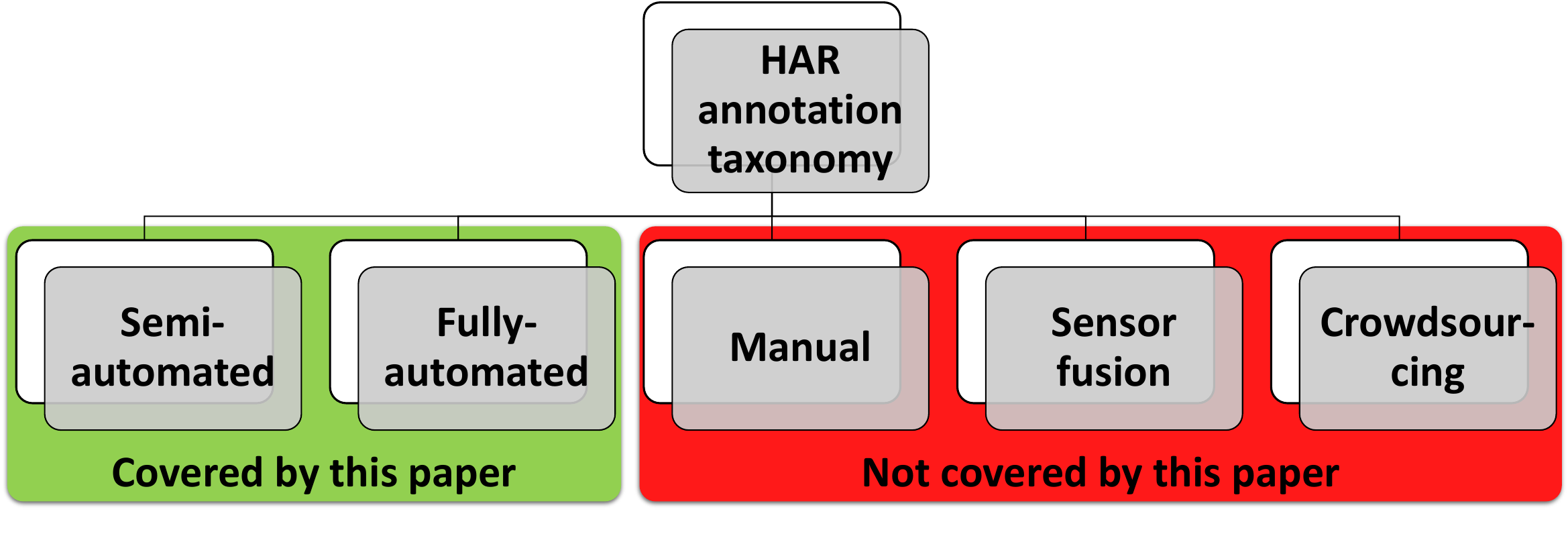}
    \caption{Overview on HAR annotation taxonomy.}
    \label{fig:tax}
\end{figure}

In this section, we provide the necessary background on data annotation techniques. 
Figure \ref{fig:tax} illustrates the different annotation techniques utilized in HAR. Each of them has unique benefits and drawbacks. This section examines and analyzes the advantages and disadvantages of these techniques. While this section provides a comprehensive overview of the technical background of annotation, Section~\ref{sec:meth} in detail describes different solutions proposed in the literature.

\subsection{Manual Annotation Systems}
Manual annotation systems require human experts to label and annotate the data manually. This approach is time-consuming, labor-intensive, and prone to errors. While manual annotation is, in principle, the golden standard and provides high-quality annotations \cite{stikic2009activity,kwapisz2011activity}, it is known to be  subjective. Hence, the results may vary between different annotators, leading to inter-annotator disagreements. The subjectivity of manual annotation can arise due to differences in annotator expertise, biases, and interpretation of the annotation guidelines. Inter-annotator disagreement can occur when multiple annotators are asked to label the same data, leading to differences in their annotations. This can reduce the reliability and validity of the annotation data, making it challenging to build machine learning models that generalize well to new, unseen data~\cite{bulling2014tutorial}. 
To mitigate these issues, manual annotation systems can incorporate various strategies, such as using multiple annotators and measuring inter-annotator agreement to ensure consistency, providing clear annotation guidelines and training to reduce subjectivity and error, and using quality control measures, such as random spot-checks and review of annotations, to ensure accuracy and completeness. Additionally, manual annotation can be supplemented with semi-automated or fully automated approaches, such as active learning, crowd-sourcing, or machine learning-assisted annotation, to increase efficiency and reduce costs.

\subsection{Semi-Automated Annotation Systems}
Semi-automated annotation systems use a combination of manual and automated annotation methods. For example, a human annotator may label a small subset of the data, and an algorithm can propagate those annotations to the rest of the dataset \cite{kwapisz2011activity, stikic2009activity}. This approach can speed up the annotation process while maintaining high-quality annotations. Semi-automated annotation systems can also reduce inter-annotator disagreement \cite{bulling2014tutorial} and can provide a middle ground between fully manual and fully automated approaches. By combining the strengths of both approaches, they can offer a more efficient and cost-effective solution for annotation tasks. 
Active learning and Transfer learning have emerged as highly innovative and accurate solutions among the semi-automated techniques.

\subsubsection{Active learning (AL)} To further improve the performance of semi-automated annotation systems, AL algorithms are designed to incorporate feedback from human annotators. For instance, an algorithm can present the most uncertain instances for annotation to human annotators, allowing them to correct errors and improve the overall quality of the labeled data. This process can reduce the number of instances that need to be labeled while maintaining the annotation quality \cite{settles2009active}.

\subsubsection{Transfer learning (TL)}
Transfer learning systems leverage pre-existing annotated datasets to train models that can be applied to new datasets \cite{cook2013transfer}. Such systems can reduce the annotation effort required and improve the accuracy of HAR algorithms, especially for similar activities across different datasets \cite{kwapisz2011activity}. 
Transfer learning can be particularly advantageous when no annotated data exists for a specific task or activity. By leveraging pre-existing annotated datasets, transfer learning annotation systems can effectively "transfer" knowledge from one dataset to another, allowing models to learn from the annotated data in one dataset and generalize to new datasets with similar activities. 
However, transfer learning annotation systems also have their own challenges, such as the need to identify appropriate pre-existing datasets that are relevant to the new dataset, and the need to carefully tune the transfer learning approach to ensure optimal performance.

\subsection{Automated Annotation Systems}
Automated annotation systems are commonly used in large-scale HAR applications, where manual annotation is not feasible due to a large amount of data \cite{bulling2014tutorial}. 
In such applications, automated annotation can help to provide a baseline for labeling the data, which can then be refined by human experts or through semi-automated methods. Such systems are fast and efficient, but their accuracy may be lower than manual or semi-automated systems, especially for complex activities, and they may require significant computational resources to train and execute \cite{kwapisz2011activity,stikic2009activity}.
Various techniques can be employed to improve the accuracy of automated annotation systems, such as feature selection and engineering, model selection, and the optimization of hyperparameters. Moreover, automated annotation techniques can be enhanced by leveraging additional sources of information, such as sensor fusion, context awareness, and domain-specific knowledge. Additionally, manual or semi-automated methods can be used to correct errors or refine the annotations produced by automatic systems.

\subsection{Sensor Fusion Annotation Systems}
Sensor fusion annotation systems combine data from multiple sensors to provide more accurate annotations. For example, combining data from accelerometers, gyroscopes, and magnetometers can give a more comprehensive picture of the user's movements \cite{bulling2014tutorial,zhang2012survey}. Sensor fusion annotation systems can improve the accuracy of HAR algorithms, especially for complex activities that are difficult to annotate with a single sensor \cite{kwapisz2011activity}. 
Sensor fusion annotation systems can also help in overcoming some of the limitations of individual sensors, such as their sensitivity to environmental factors or their limited coverage of certain types of movements. 
However, sensor fusion annotation systems also have their own challenges, such as the need for careful calibration and synchronization of multiple sensors, and the complexity of combining data from different sources. Moreover, the increased amount of data generated by sensor fusion systems can require more powerful computational resources and more sophisticated algorithms to process and analyze.

\subsection{Crowdsourcing Annotation Systems}
Crowdsourcing annotation systems use crowdsourcing platforms to collect annotations from a large pool of non-expert annotators \cite{capponi2019survey,do2011crowdsourcing, yu2012crowdsourcing}. Crowdsourcing can provide access to a diverse pool of annotators, allowing for annotations to be collected from a range of perspectives and backgrounds. 
This approach can be cost-effective and scalable, but the quality of the annotations may vary depending on the expertise and motivation of the crowd workers. Such systems can also introduce noise and errors in the annotations, which may require additional quality control measures~\cite{bulling2014tutorial, kwapisz2011activity}, such as redundant annotations or expert reviews. 
Moreover, crowdsourcing annotation systems can introduce challenges related to task design and management, such as the need to design effective annotation tasks that are understandable and accessible by non-expert annotators, and the need to manage and monitor the crowd workers to ensure that high-quality annotations are collected.\\

\noindent In summary, the choice of an annotation system for HAR depends on various factors, such as the availability of annotated data, the complexity of the activities to be annotated, the size of the dataset, and the resources available. Every annotation technique, as summarized in Tables~\ref{tab:annot} and~\ref{tab:annot_1}, has advantages and disadvantages, and researchers must carefully evaluate which approach is most suitable for their specific HAR task.

\begin{table*}[!bh]
\centering
\caption{Characteristics of Data Annotation Techniques Used in HAR}\label{tab:annot}
\resizebox{\textwidth}{!}{
\begin{tabular}{|c||c|c|c|c|c|c|c|c|}
\hline
\textbf{Technique} & 
\makecell{\textbf{Expertise}\\\textbf{Level}}&
\makecell{\textbf{Annotation}\\\textbf{Complexity}}&
\makecell{\textbf{Time}\\\textbf{Consumption}}&
\makecell{\textbf{Economic} \\\textbf{Costs}}&
\makecell{\textbf{Annotation}\\\textbf{Accuracy}}&
\makecell{\textbf{Inter-annotator}\\ \textbf{Agreement}}&
\textbf{Scalability}&
\textbf{Subjectivity}\\ \hline\hline
Manual & High&High&High & High & High & Moderate & Low & High \\ \hline
\makecell{Semi-\\automatic} & Moderate-High& Moderate & Moderate & Moderate & High & High & High & High \\ \hline
Automatic & High&Low & Low & Low & \makecell{Low-\\Moderate} & \makecell{Low-\\Moderate} & \makecell{Low-\\Moderate} & Moderate \\ \hline
\makecell{Sensor\\ Fusion} & High&High & High & High & High & High & High & \makecell{Low-\\Moderate} \\ \hline
\makecell{Crowd-\\sourcing} & \makecell{Low-\\Moderate} & \makecell{Low-\\Moderate} & \makecell{Low-\\Moderate} & \makecell{Low-\\Moderate} & \makecell{Low-\\Moderate} & \makecell{Low-\\Moderate} & \makecell{Low-\\Moderate} & High \\ 
\hline\hline
\end{tabular}}
\end{table*}

\begin{table*}[!bh]
\centering
\caption{Advantages and Limitations of Data Annotation Techniques Used in HAR}\label{tab:annot_1}
\resizebox{\textwidth}{!}{
\begin{tabular}{|c||c|c|c|}
\hline
\textbf{Technique} & 
\textbf{Requirements} & 
\textbf{Advantages} & 
\textbf{Limitations} \\ \hline\hline
Manual & 
\makecell{Human annotators with expertise,\\ time and resources for manual annotation} & 
\makecell{High accuracy, ability to\\ handle complex activities} & 
\makecell{Time-consuming, labor-intensive,\\ subjective, requires human expertise} \\ \hline

\makecell{Semi-\\automatic} & 
\makecell{Annotated training \\data, human annotators, \\and appropriate algorithms} &
\makecell{Faster than manual annotation,\\ maintains high-quality annotations,\\ ability to reduce human errors} & 
\makecell{Requires significant\\ human involvement, \\not fully automated} \\ \hline

Automatic &
\makecell{Dedicated systems \\and/or algorithms} &
\makecell{Fast, scalable, can\\ handle large datasets} & 
\makecell{Lower accuracy compared to\\ manual or semi-automated systems,\\ requires high-quality sensor data, limited\\ ability to handle complex activities} \\ \hline

\makecell{Sensor\\ Fusion} & 
\makecell{Multiple sensors capturing\\ relevant information,\\ integration of sensor data} &
\makecell{Enhances more accurate\\ annotations by combining data \\from multiple sensors} & 
\makecell{Requires manual annotation\\ and additional sensor data,\\ may be computationally intensive} \\ \hline

\makecell{Crowd-\\sourcing} & 
\makecell{Crowdsourcing platform, large\\ pool of non-expert annotators,\\ quality control measures} &
\makecell{Can collect annotations from\\ a large pool of individuals,\\ reduces cost and time} & 
\makecell{Lower accuracy compared to\\ expert annotators, requires careful\\ selection and training of crowd workers } \\ \hline\hline
\end{tabular}}
\end{table*}

%% file: sec/3-search.tex

\section{Selection criteria}\label{sec:search}
This section describes the selection criteria of this systematic review, i.e., how the papers that were considered were selected. 
This review includes only studies focused on developing and evaluating (semi-, fully-) automated data annotation techniques for HAR. The participants were required to be human,  while studies involving non-human subjects were excluded. In addition, studies had to report on the accuracy, precision, and other relevant performance metrics of the annotation systems. 
Only publications in English language were considered, and all studies had to be published in peer-reviewed journals or conference proceedings.
The search strategy and selection criteria were developed in consultation with all authors. Any disagreements between reviewers were resolved through discussion and consensus. The study selection process was documented using a Preferred Reporting Items for Systematic Reviews and Meta-Analyses (PRISMA) flowchart to ensure transparency and applicability \cite{moher2009preferred}.\\
\begin{table}[!ht]
\centering
\caption{Executed search query on IEEE Explore, the ACM digital libary, Scopus, and Web of Science on 21/01/2023.}\label{tab:query}
\resizebox{0.9\textwidth}{!}{
\begin{tabular}{|p{.975\textwidth}|c|}
\hline
\hline
active learning \textbf{or} semi supervised learning \textbf{or} semi-supervised learning \textbf{or} activity recognition \textbf{or} human activity recognition \textbf{or} HAR \textbf{or} pattern recognition \textbf{or} machine learning \textbf{or} deep learning & \textbf{and} \\
\hline
automatic data label(l)ing \textbf{or} semi-automatic data label(l)ing \textbf{or} automated data label(l)ing \textbf{or} semi-automated data label(l)ing \textbf{or} data annotation \textbf{or} automatic data annotation \textbf{or} semi-automatic data annotation \textbf{or} semi-automated data annotation
%
 & \textbf{and} \\
 \hline
 wearable \textbf{or} wearables \textbf{or} wearable sensors \textbf{or} body-worn sensors \textbf{or} inertial sensor \textbf{or} inertial measurement unit \textbf{or} video \textbf{or} smartphone  \textbf{or} smartwatch \textbf{or} smart glasses        & \\
 \hline
 \hline
\end{tabular}}
\end{table}

\noindent \textbf{PRISMA Flowchart:} Table~\ref{tab:query} presents the search query used to identify relevant studies during the research phase of the systematic review. The query was structured into three categories or "leaves" that represent the main concepts of interest in the review: 1) algorithms, 2) automated annotation systems, and 3) devices. These three concepts form the basis of the inclusion criteria for selecting studies considered by the systematic review. By specifying the types of algorithms, annotation processes, and devices of interest, the query helps to ensure that the studies selected for the review are relevant and meet the specific research objectives. \\

Table \ref{tab:serch_outcome} shows the number of search results retrieved from each of the four databases (i.e., IEEE Explore, ACM Digital Library, Scopus, and Web of Science) on January 21, 2023), using the search strategy defined for the systematic review.
\begin{table}[!ht]
\centering
\caption{Search outcome samples per Publisher from 01/01/1980 to 21/01/2023.}\label{tab:serch_outcome}
 
 
\begin{tabular}{|c||cccc|}
\hline
 \textbf{Publisher} & IEEE Explore & ACM Digital Library & Scopus & Web of Science \\ \hline\hline
 \makecell{\textbf{Link to search query}}  &  \href{https://ieeexplore.ieee.org/search/searchresult.jsp?action=search&matchBoolean=true&newsearch=true&queryText=((\%22Active\%20learning\%22\%20OR\%20\%22Semi\%20supervised\%20learning\%22\%20OR\%20\%22Semi-supervised\%20learning\%22\%20OR\%20\%22Activity\%20Recognition\%22\%20OR\%20\%22Human\%20Activity\%20Recognition\%22\%20OR\%20\%22HAR\%22\%20OR\%22pattern\%20recognition\%22\%20OR\%20\%22machine\%20learning\%22\%20OR\%20\%22deep\%20learning\%22)\%20AND\%20(\%22automatic\%20data\%20labelling\%22\%20OR\%20\%22semi-automatic\%20data\%20labelling\%22\%20OR\%20\%22automatic\%20labelling\%22\%20OR\%20\%22semi-automatic\%20labelling\%22\%20OR\%20\%22data\%20annotation\%22\%20OR\%20\%22automatic\%20data\%20annotation\%22\%20OR\%20\%22semi-automatic\%20data\%20annotation\%22\%20OR\%20\%22semi\%20automatic\%20data\%20annotation\%22\%20OR\%20\%22automated\%20data\%20labelling\%22\%20OR\%20\%22semi-automated\%20data\%20labelling\%22\%20OR\%20\%22automated\%20labelling\%22\%20OR\%20\%22semi-automated\%20labelling\%22\%20OR\%20\%22data\%20annotation\%22\%20OR\%20\%22automatic\%20data\%20annotation\%22\%20OR\%20\%22semi-automatic\%20data\%20annotation\%22\%20OR\%20\%22semi\%20automatic\%20data\%20annotation\%22\%20OR\%20\%22automatic\%20data\%20labeling\%22\%20OR\%20\%22semi-automatic\%20data\%20labeling\%22\%20OR\%20\%22automatic\%20labeling\%22\%20OR\%20\%22semi-automatic\%20labeling\%22\%20OR\%20\%22data\%20annotation\%22\%20OR\%20\%22automatic\%20data\%20annotation\%22\%20OR\%20\%22semi-automatic\%20data\%20annotation\%22\%20OR\%20\%22semi\%20automatic\%20data\%20annotation\%22\%20OR\%20\%22automated\%20data\%20labeling\%22\%20OR\%20\%22semi-automated\%20data\%20labeling\%22\%20OR\%20\%22automated\%20labeling\%22\%20OR\%20\%22semi-automated\%20labeling\%22\%20OR\%20\%22data\%20annotation\%22\%20OR\%20\%22automatic\%20data\%20annotation\%22\%20OR\%20\%22semi-automatic\%20data\%20annotation\%22\%20OR\%20\%22semi\%20automatic\%20data\%20annotation\%22)\%20AND\%20(\%22wearable\%22\%20OR\%20\%22wearables\%22\%20OR\%20\%22wearable\%20sensors\%22\%20OR\%20\%22body-worn\%20sensors\%22\%20OR\%20\%22inertial\%20sensor\%22\%20OR\%20\%22inertial\%20measurement\%20unit\%22\%20OR\%20\%22video\%22\%20OR\%20\%22smartphone\%22,\%20OR\%20\%22smartwatch\%22\%20OR\%20\%22smart\%20glasses\%22))}{link}      & \href{https://dl.acm.org/action/doSearch?fillQuickSearch=false&target=advanced&expand=dl&AllField=AllField\%3A\%28\%28\%22Active+learning\%22+OR+\%22Semi+supervised+learning\%22+OR+\%22Semi-supervised+learning\%22+OR+\%22Activity+Recognition\%22+OR+\%22Human+Activity+Recognition\%22+OR+\%22HAR\%22+OR\%22pattern+recognition\%22+OR+\%22machine+learning\%22+OR+\%22deep+learning\%22\%29+AND+\%28\%22automatic+data+labelling\%22+OR+\%22semi-automatic+data+labelling\%22+OR+\%22automatic+labelling\%22+OR+\%22semi-automatic+labelling\%22+OR+\%22data+annotation\%22+OR+\%22automatic+data+annotation\%22+OR+\%22semi-automatic+data+annotation\%22+OR+\%22semi+automatic+data+annotation\%22+OR+\%22automated+data+labelling\%22+OR+\%22semi-automated+data+labelling\%22+OR+\%22automated+labelling\%22+OR+\%22semi-automated+labelling\%22+OR+\%22data+annotation\%22+OR+\%22automatic+data+annotation\%22+OR+\%22semi-automatic+data+annotation\%22+OR+\%22semi+automatic+data+annotation\%22+OR+\%22automatic+data+labeling\%22+OR+\%22semi-automatic+data+labeling\%22+OR+\%22automatic+labeling\%22+OR+\%22semi-automatic+labeling\%22+OR+\%22data+annotation\%22+OR+\%22automatic+data+annotation\%22+OR+\%22semi-automatic+data+annotation\%22+OR+\%22semi+automatic+data+annotation\%22+OR+\%22automated+data+labeling\%22+OR+\%22semi-automated+data+labeling\%22+OR+\%22automated+labeling\%22+OR+\%22semi-automated+labeling\%22+OR+\%22data+annotation\%22+OR+\%22automatic+data+annotation\%22+OR+\%22semi-automatic+data+annotation\%22+OR+\%22semi+automatic+data+annotation\%22\%29+AND+\%28\%22wearable\%22+OR+\%22wearables\%22+OR+\%22wearable+sensors\%22+OR+\%22body-worn+sensors\%22+OR+\%22inertial+sensor\%22+OR+\%22inertial+measurement+unit\%22+OR+\%22video\%22+OR+\%E2\%80\%9Csmartphone\%E2\%80\%9D\%2C+OR+\%E2\%80\%9Csmartwatch\%E2\%80\%9D+OR+\%E2\%80\%9Csmart+glasses\%E2\%80\%9D\%29\%29}{link} 
 & \href{https://www.scopus.com/results/results.uri?sort=plf-f&src=s&sid=10fd2c7b443c4e25c35c470f9a0b9959&sot=a&sdt=a&sessionSearchId=10fd2c7b443c4e25c35c470f9a0b9959&origin=searchadvanced&editSaveSearch=&txGid=8851693530e0274079f0389b35f9abc6}{link} & \href{https://www.webofscience.com/wos/woscc/summary/c4507b9b-7a43-48d4-9705-10495079bd1b-21b9bb47/relevance/1}{link} \\ \hline
Total & 64 & 611 & 1650 & 76\\
 \hline
 \hline
\end{tabular}
\end{table}

\begin{figure}[!ht]
    \centering
    \includegraphics[width=.625\textwidth,page={3}]{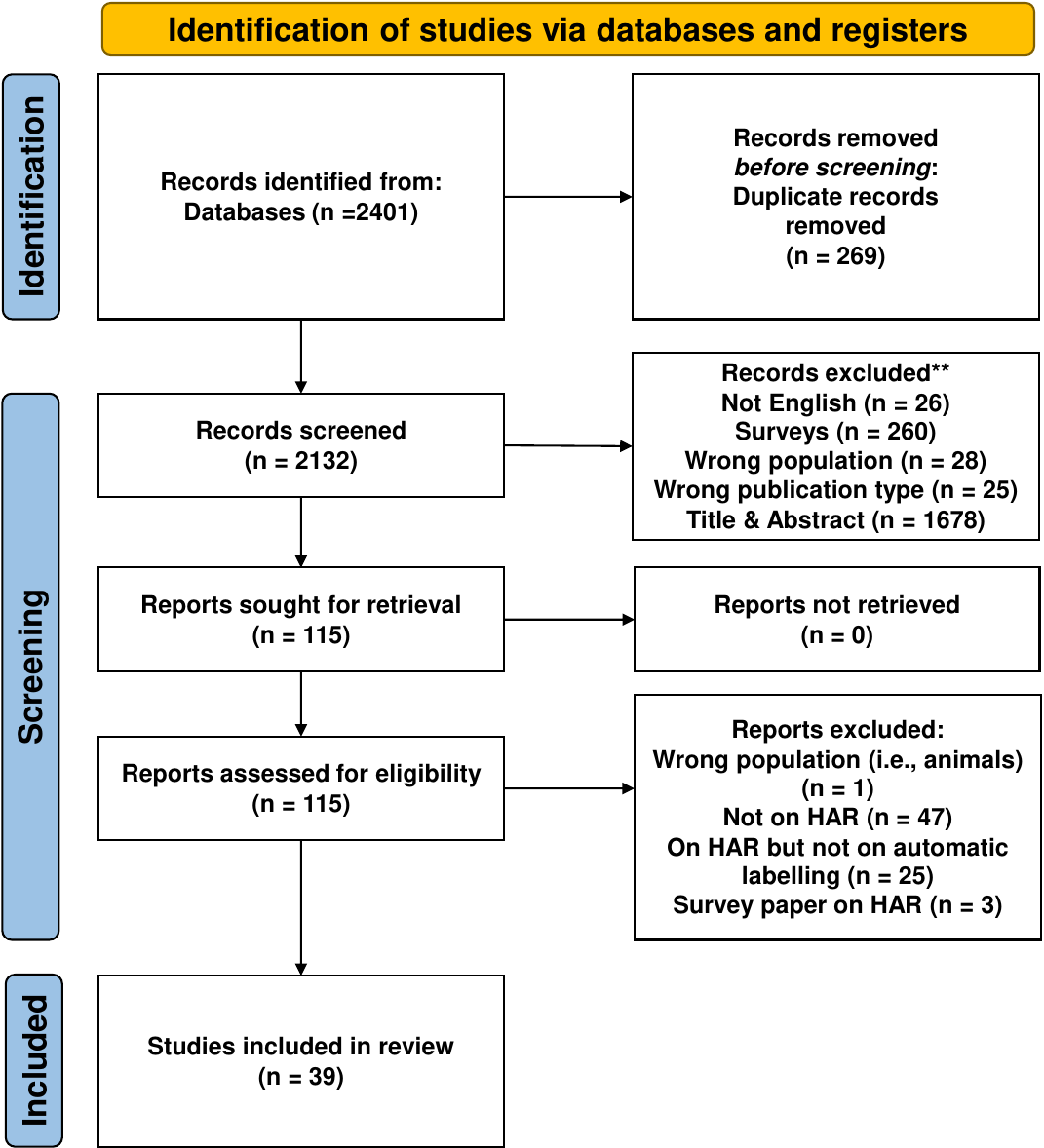}
    \caption{PRISMA diagram of the performed search.}
    \label{fig:prisma}
\end{figure}

Figure \ref{fig:prisma} illustrates the PRISMA flowchart, which serves as a transparent and replicable means of reporting the systematic review's search and selection process. 

The chart shows that 2401 research articles were initially retrieved through the search process described in Table \ref{tab:query}. 
It then depicts the screening process, ultimately leading to the inclusion of 39 studies in the review. 
Excluded research items did not meet the pre-defined selection criteria outlined at the beginning of this section.

Table~\ref{tab:dist} illustrates the distribution of the 115 research items assessed for eligibility over time and reveals a growing interest in the field of HAR technologies and automatic data annotation techniques. 
The table provides information on the number of studies published in each year, distinguishing between included and excluded items, and offers a glimpse into the research activity in this field over time. 
Notably, the table shows that despite the search starting as early as 01/01/1980, no work in this field was presented until 2006. 
By showcasing the increasing number of studies on HAR technologies and automated data annotation, the Table~\ref{tab:dist} implies that this subject is gaining more prominence and significance in the field, providing a comprehensive overview of the literature landscape that can assist researchers in identifying trends, gaps, and areas for further exploration.\\

\begin{table}[!ht]
\centering
\caption{Distribution of selected papers over time.}\label{tab:dist}
\resizebox{0.9\textwidth}{!}{
\begin{tabular}{c||cccccccccccccccccccc||c}
\hline
Year (20xx)& 04&05&06&07&08&09&10&11&12&13&14&15&16&17&18&19&20&21&22&23&Total\\ \hline\hline
Included & 0&0&1&0&0&1&0&0&0&1&0&2&2& 3& 6&8&3& 7& 5& 0& 39 \\
Excluded & 1&1&4&1&0&0&0&2&0&2&2&3&5& 9& 6&9&4& 8&18& 1& 76 \\ \hline
Total    & 1&1&5&1&0&1&0&2&0&3&2&5&7&12&12&17&7&15&23& 1&115\\
 \hline \hline
\end{tabular}
}
\end{table}

Finally, out of the 2401 reviewed papers, none of them were found to be survey papers on data annotation techniques in HAR. As a result, we claim that this is the first systematic review to address this topic. 

%% file: sec/4-meth.tex
\section{Annotation Systems in HAR}\label{sec:meth}
Based on the analysis of the 39 out of 2401 papers identified through our selection criteria procedure, this systematic review will focus solely on semi-automated and fully-automated data annotation techniques and exclude manual techniques, sensor fusion, and crowdsensing. 
Thus, the techniques will be categorized into semi- and fully-automated and subsequently into three categories: a) data-driven, b) environment-driven, and c) hybrid. 

\noindent \textbf{Data-driven:} Data-driven techniques leverage the patterns, structures, and characteristics inherent in the data itself to guide the annotation process.\\
\noindent \textbf{Environment-driven:} The environment-driven techniques use information about the context and environment in which the data was collected to perform annotation. For example, use the interaction of users with ODLs to recognize the performed activity. \\
\noindent \textbf{Hybrid:} The hybrid techniques combine both data-driven and environment-driven approaches, often using multiple sources of information to achieve more accurate and robust annotation results.\\ 

\begin{figure}[!ht]
    \centering
    \includegraphics[width=0.875\textwidth,page={4}]{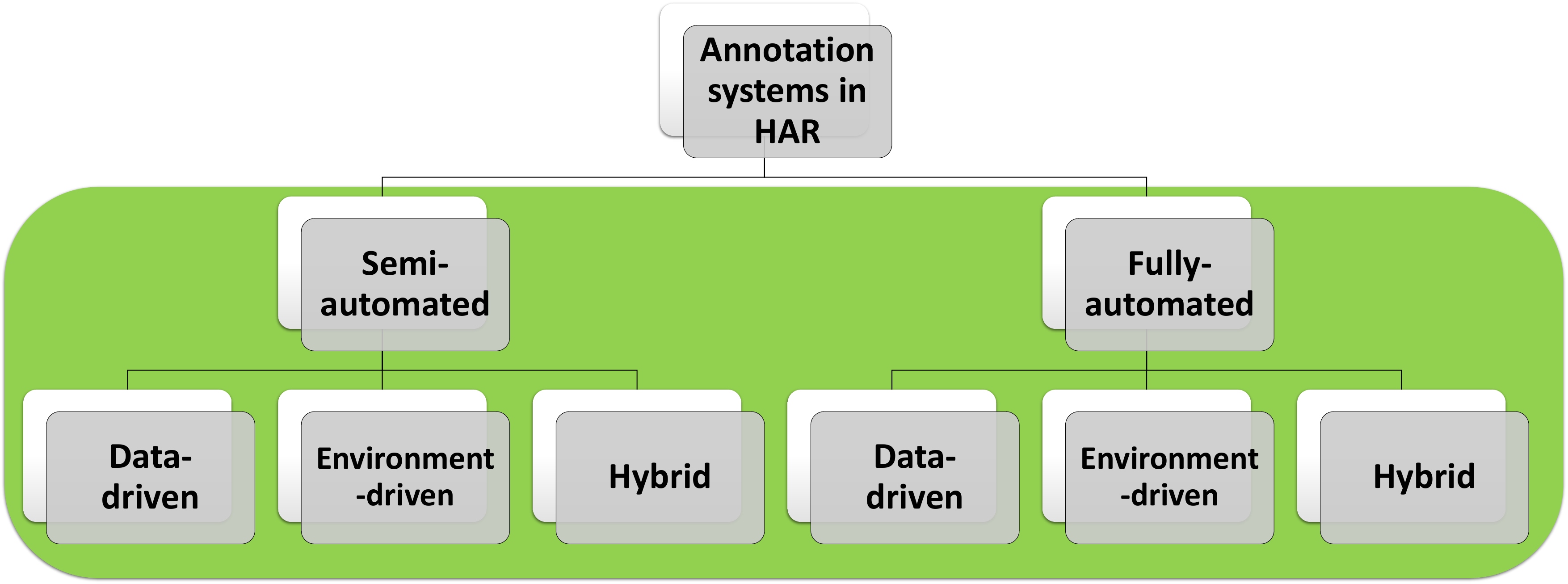}
    \caption{Automatic data annotation taxonomy in HAR.}
    \label{fig:tax_2}
\end{figure}

This taxonomy is shown in Figure \ref{fig:tax_2}, while Table \ref{tab:all-papers} provides an overview of the studies included in the review and categorized using the above taxonomy.

\begin{table*}[!ht]
\centering
\caption{Overview of studies included in this reviewed paper grouped by the mentioned taxonomy (\cite{szewcyzk2009annotating} fits into two categories).}\label{tab:all-papers}
\resizebox{0.875\textwidth}{!}{
\begin{tabular}{|c|c|c|c|c|c|}
\hline
\multicolumn{6}{|c|}{\textbf{Data annotation techniques in HAR}}\\ \hline \hline
\multicolumn{3}{|c|}{\textbf{Semi-automated}} &\multicolumn{3}{c|}{\textbf{Fully-automated}} \\ \hline
\textbf{Data-driven} &\textbf{Environment-driven}& \textbf{Hybrid} &\textbf{Data-driven}&\textbf{Environment-driven}&\textbf{Hybrid}\\
\hline \hline
\makecell{
\cite{saeedi2017co} \cite{martindale2018smart} \cite{gan2018automatic} \cite{bota2019semi} \\ \cite{martindale2019hidden} \cite{ponnada2019designing} \cite{faridee2019augtoact} \cite{hossain2019active}\\ \cite{sawano2020annotation} \cite{kwon2020imutube} \cite{avsar2021benchmarking} \cite{korpela2021reducing} \\ \cite{tang2021selfhar} \cite{yhdego2022fall} \cite{mohamed2022har}}&

\makecell{\cite{szewcyzk2009annotating} \cite{subramanya2012recognizing} \cite{woznowski2017talk} \cite{tonkin2018talk}\\ \cite{solis2019human}}&

\makecell{\cite{alam2015mobeacon} \cite{meurisch2015labels} \cite{nino2016scalable} \cite{cruciani2018personalized}\\ \cite{cruz2019semi} \cite{tonkin2019towards}}&

\makecell{\cite{jardim2016automatic} \cite{rokni2018autonomous} \cite{liang2018automatic} \cite{zhang2020deep} \\\cite{jeng2021wrist} \cite{ullrich2021detection} \cite{ma2021unsupervised} \cite{qi2022dcnn}\\\cite{lin2022clustering}}&

\makecell{\cite{szewcyzk2009annotating} \cite{loseto2013mining} \cite{al2017annotation} \cite{demrozi2021towards} \\ \cite{dissanayake2021indolabel}}&
\\ \hline \hline
Total: 15&Total: 5*&Total: 6&Total: 9&Total: 5*&Total: 0\\
\hline
\end{tabular}}
\end{table*}

In the order of their date of publication, this section will provide a comprehensive overview of each paper included in this survey. Additionally, for each category of our taxonomy, a table that summarizes the most important properties of all publications that fall under the corresponding category is provided.
In particular, Tables \ref{tab:S-DD}, \ref{tab:S-ED}, \ref{tab:S-H}, \ref{tab:H-DD}, and \ref{tab:H-ED} contain detailed information on aspects such as the ownership (i.e., authors used an open access existing dataset or designed and collected new data on their own) of the tested datasets and the number of datasets utilized (Column 2), the devices (Column 3) and sensors (Column 4) employed, the total number of used sensors (Column 5), their placement on the body or in the environment (Column 6), the ML/DL/AI models utilized (Column 7), the number of subjects involved in data collection (Column 8), the performed activities (Column 9), the total number of activities (Column 10), and the type of considered daily environment (Column 11).

Instead, Tables \ref{tab:S-DD-1}, \ref{tab:S-ED-1}, \ref{tab:S-H-1}, \ref{tab:H-DD-1}, and \ref{tab:H-ED-1} provide the annotation techniques used by the individual publications, and summarizes the individual advantages (Column 3) and disadvantages (Column 4) of every approach.

Moreover, in order to improve the legibility of Tables \ref{tab:S-DD} through \ref{tab:H-ED-1}, Table \ref{tab:glossary} offers a summary of the terminology utilized to classify the different types of activities discussed in the methodologies of the examined papers. 
\begin{table}[!ht]
\centering
\caption{Glossary of abbreviations used to categorize human activities/status types.}\label{tab:glossary}
\resizebox{0.72\textwidth}{!}{
\begin{tabular}{c|p{2cm}|p{8cm}}
\hline
\textbf{Acronym} & \textbf{Extended} & \textbf{Definition}\\
\hline\hline
\textbf{BHA} & Basic Human Activities & The class of activities comprising \emph{walking}, \emph{running}, and \emph{sleeping} can be broadly defined as ``basic human activities''. These activities are fundamental to human life and are integral parts of our daily routines.\\
\hline
\textbf{OBI} & Object-based Interactions & The class of activities comprising interactions with objects such as \emph{smartphones} or \emph{fridges} can be referred to as ``object-based interactions''. These activities involve engaging with specific objects to perform various tasks or actions.\\
\hline
\textbf{ADL} & Activities of Daily Living &  ADLs are a set of essential everyday activities that individuals typically perform as part of their daily routines to maintain their personal well-being and independence. These can be: personal hygiene, eating, or cooking.\\
\hline
\textbf{PA} & Protocol Activities & Human activities within an experimental protocol refer to activities that individuals were asked to carry out solely for the study. They can vary depending on the specific nature of the experiment and the research objectives.\\
\hline
\textbf{R} & Risky & Risky activities refer to actions or behaviors that involve an increased potential for accidents, injuries, or negative outcomes. One example of a risky activity is falling.\\
\hline\hline
\end{tabular}}
\end{table}

Conversely, Table \ref{tab:abbreviations} presents an overview of the acronyms used for the annotation models.

\begin{table}[!ht]
\centering
\caption{List of acronyms used throughout the paper}\label{tab:abbreviations}
\resizebox{0.95\textwidth}{!}{\begin{tabular}{c|c||c|c}
\hline
\textbf{Acronym} & \textbf{Definition}&\textbf{Acronym} & \textbf{Definition} \\
\hline
\hline
AL & Active Learning & LSTM & Long Short-Term Memory\\
BEL & Bagging Ensemble Learning & MEEL & Minimum-Error Exact Labeling\\
BLE & Bluetooth Low Energy & ML & Machine Learning\\
cGAN & Conditional Generative Adversarial Network & n/a & Not Available\\
CRF & Conditional Random Field & NFC & Near-Field Communication\\
cVAE-cGAN & Conditional Variational AutoEncoder-cGAN & NN & Neural Network\\
DCNN & Deep Convolutional Neural Network & ODLs & Objects of Daily Life \\
DL & Deep Learning & PIR & Passive Infrared Sensor\\
DT & Decision Tree & PNB & Packaged Naive Bayes\\
DTW & Dynamic Time Warping & PSE & Particle Swarm Optimization\\
FCN & Fully Connected Network & QS & Query Strategies\\
FS-FCPSO & Fuzzy C-means Particle Swarm Optimization & ResNet & Residual Neural Network\\
GPS & Global Positioning System & RF & Random Forest\\
HAR & Human Activity Recognition & RSSI & Received Signal Strenght Indicator\\
HAR-GCNN & HAR-Graph Chronologically Correlation Network & SelfHAR & Self-training and Self-Supervised Learning HAR\\
HAS & Human Activity Sequence & SLIC & Simple Linear Iterative Clustering\\
HCG & Human Computation Games & SMO & Sequential Minimal Optimization\\
hHMM & Hierarchical HMM & SQ & Stopping Criteria\\
Hk-Medoids & Hierarchical k-Medoids & SSAL & Semi-Supervised AL\\
HMM & Hidden Markov Model & ST & Self-Training\\
HR & Hart Rate & SVM & Support Vector Machine\\
IMU & Inertial Measurement Unit & TL & Transfer Learning\\
k-NN & k-Nearest Neighbor &&\\
\hline
\hline
\end{tabular}}
\end{table}

\subsection{Semi-Automated Annotation}\label{sec:4.1}
Within this section, the 26\footnote{paper \cite{szewcyzk2009annotating} fits into both fully- and semi-automated categories, since it provides different approaches.\label{fo:54}} papers on semi-automated data annotation techniques for HAR are categorized into the three subcategories mentioned earlier, and their proposed methods and key features are described in detail.

\subsubsection{Data-driven approaches}\label{sec:4.1.1}
These studies propose semi-automated data-driven methodologies for automated data annotation in HAR, leveraging the existence of data patterns being extracted through various techniques such as AL, augmented and TL frameworks, or self-supervised learning. 
Tables \ref{tab:S-DD} and \ref{tab:S-DD-1} offer a comprehensive summary of the 15 identified works falling in this category.

\begin{table*}[!th]
\centering
\caption{Semi-automated Data-Driven approaches.}\label{tab:S-DD}
\resizebox{\textwidth}{!}{
\begin{tabular}{c||cccccccccc}
\hline
\makecell{\textbf{Ref.}\\ \textbf{(Year)}} & \makecell{\textbf{Dataset}\\\textbf{Provided (\#)}} & \textbf{Devices} & \textbf{Sensors} & \makecell{\textbf{\# of}\\ \textbf{Sensors}} & \textbf{Position} & \textbf{Model} & \makecell{\textbf{\# of} \\\textbf{Subjects [M/F]}} & \textbf{Activities} & \makecell{\textbf{\# of} \\\textbf{Activities}} & \makecell{\textbf{In/Out}\\ \textbf{Door}}\\ \hline \hline
\makecell{\cite{saeedi2017co}\\ (2017)}
&no (1)
&IMU
&\makecell{accelerometer,\\ gyroscope}
&7
&\makecell{chest, shin, forearm,\\ thigh, upper-arm, head,\\ and waist}
&\makecell{k-NN\\ DT\\ RF}
&15 [8/7]
&\makecell{BHA}

&8
&indoor\\ \hline

\makecell{\cite{martindale2018smart}\\ (2018)}
&yes (1)
&\makecell{IMU, video,\\ pressure\\ insoles}
&\makecell{accelerometer,\\ gyroscope,\\ Pressure}
&7
&\makecell{trouser pocket, wrists, shoe\\ side, under-feet}
&Threshold
&20 [15/5]
&\makecell{BHA, PA}
&13
&\makecell{indoor\\ outdoor}\\ \hline

\makecell{\cite{gan2018automatic}\\ (2018)}
&yes (1)
&\makecell{smartphone,\\ smartwatch}
&accelerometer
&2
&wrist
&\makecell{SVM, RF,\\ K-NN, NN}
&1
& R
&1
&indoor\\ \hline

\makecell{\cite{bota2019semi}\\ (2019)}
&no (2)
&IMU
&\makecell{accelerometer,\\ gyroscope}
&1-4
&\makecell{waist - right/left hand,\\ right ankle, and right\\ side of the waist}
&SSAL
&30 - 12
&\makecell{BHA}
&6 - 7
&indoor\\ \hline

\makecell{\cite{martindale2019hidden}\\ (2019)}
&yes (1)
&IMU
&\makecell{accelerometer,\\ gyroscope}
&5
&\makecell{taurus, wrists, shoe\\ side, under-feet}
& HMM
&80 [52/28]
&\makecell{BHA, PA, OBI}
&12
&indoor\\ \hline

\makecell{\cite{ponnada2019designing}\\ (2019)}
&yes (1)
&actigraph
&accelerometer
&1
&wrist
&\makecell{Mobots and\\ Signaligner games}
&50
&\makecell{BHA, OBI}
&6
&indoor\\ \hline

\makecell{\cite{faridee2019augtoact}\\ (2019)}
&no (2)
&\makecell{smartphone,\\ smartwatch}
&accelerometer
&8-4 \textit{n/a}
&\makecell{chest, head, shin,\\ thigh, upper arm waist - \textit{n/a}}
&AugToAct
&15 / NA
&\makecell{BHA}
&6 - 10
&\makecell{indoor\\ outdoor}\\ \hline

\makecell{\cite{hossain2019active}\\ (2019)}
&yes (1)
&smartphone
&accelerometer
&1
&front pant pocket
&\makecell{Deep-AL and\\ Actor-Critic Network}
&20 [14/6]
&\makecell{BHA}
&6
&\makecell{indoor\\ outdoor}\\ \hline

\makecell{\cite{sawano2020annotation}\\ (2020)}
&yes (1)
&smartphone
&accelerometer
&1
&\makecell{hand, on a table, pocket,\\ walking, standing }
&\makecell{Semi\\ Supervised\\ Learning}
&5
&\makecell{OBI}
&10
&indoor\\ \hline 

\makecell{\cite{kwon2020imutube}\\ (2020)}
&no (11)
&\makecell{virtual\\ IMU}
&\makecell{accelerometer,\\ gyroscope}
&11-3
&\makecell{left and right\\ feet, left shin and\\ thigh, hip, back, left\\ and right arms, left \\and right forearms - \textit{n/a}}
&\makecell{DeepConv\\ LSTM}
&9 - 14
&\makecell{BHA}
&11 - 8
&\makecell{indoor\\ outdoor}\\ \hline

\makecell{\cite{avsar2021benchmarking}\\ (2021)}
&no (1)
&video
&\textit{n/a}
&\textit{n/a}
&\textit{n/a}
&\makecell{Temporal\\ Convolutional\\ Neural-Network}
&14
&\makecell{BHA, PA}
&8
&indoor\\ \hline

\makecell{\cite{korpela2021reducing}\\ (2021)}
&no (4)
&video
&\textit{n/a}
&\textit{n/a}
&\textit{n/a}
&\makecell{Simple Linear\\ Iterative Clustering and\\ Image Segmentation}
&\textit{n/a}
&\makecell{BHA, ADL, OBI}
&12 - 18 - 5 - 5
&indoor\\ \hline

\makecell{\cite{tang2021selfhar}\\ (2021)}
&no (8)
&IMU
&\makecell{accelerometer,\\ gyroscope}
&\makecell{2 - 1 - 1 \\- 1 - 1 - 1\\ - 1 - 1}
&\makecell{waist, arm - front pocket\\ - front pocket - wrist\\ - front pocket - waist\\ - front pocket - wrist}
&SelfHAR
&\makecell{9 - 24 - 66\\ - 28 - 30 - 30\\ - 29 - 2096}
&\makecell{BHA, ADL, R}
&\makecell{6 - 6 - 11\\ - 11 - 9 - 6\\ - 6 - \textit{n/a}}
&\makecell{indoor\\ outdoor}\\ \hline 

\makecell{\cite{yhdego2022fall}\\ (2022)}
&yes (1)
&VR headset
&\makecell{accelerometer,\\ gyroscope}
&1
&shank
&ResNet
&16 [14/2]
& R
&1
&indoor\\ \hline

\makecell{\cite{mohamed2022har}\\ (2022)}
&no (2)
&smartwatch
&IMU, HR
&3-4
& \textit{n/a} - \textit{n/a}
&GCNNs
&60 - 9
&\makecell{BHA}
&51 - 12 
&indoor\\ \hline \hline
\multicolumn{11}{c}{In Columns Devices, \# of Sensors, Position, \# of Subjects [M/F], Activities, and \# of Activities the symbol \"-\" is used to separate information of the different datasets.}
\end{tabular}}
\end{table*}

\begin{table*}[!th]
\centering
\caption{Annotation methods, Advantages, and Limitations of Semi-automated Data-Driven approaches.}\label{tab:S-DD-1}
\resizebox{\textwidth}{!}{
\begin{tabular}{c||p{7cm}p{5cm}p{5cm}}
\hline
\textbf{Ref. (Year)} & \textbf{Annotation} & \textbf{Advantages} & \textbf{Limitations} \\ \hline \hline
\makecell{\cite{saeedi2017co}\\ (2017)}
& TL/AL
& The accuracy of activity recognition reaches over 85\% by labeling only 15\% of unlabeled data.
& Requires 15\% of labeled data.\\ \hline

\makecell{\cite{martindale2018smart}\\ (2018)}
& Four methods: 1) EdgeDet1: Detection of rising and falling edges of individual pressure sensors.\newline
2) EdgeDet2: Detection of rising and falling edges of all 5 pressure sensors’ profiles.\newline
3) Threshold: De-drift using envelope subtraction, followed by detection of threshold crossings.\newline
4) Moticon: Moticon algorithm based on 100 Hz data.
& Uses walking cycle phase recognition to identify activities.
& Requires feedback for label validation, and 17\% manual labeling or correction is needed.\\ \hline

\makecell{\cite{gan2018automatic}\\ (2018)}
& Threshold-based and Clustering
& Achieves up to 97\% fall detection accuracy and at least 96\% step counting accuracy with automatic labeling of peak-trough magnitudes.
& Requires observer feedback to identify some fall contexts.\\ \hline

\makecell{\cite{bota2019semi}\\ (2019)}
&SSAL
&Reduces the required annotated data by more than 89\%.
&Requires labeled data.\\ \hline

\makecell{\cite{martindale2019hidden}\\ (2019)}
& Annotates data from 20 subjects, trains an hHMM, uses the model to annotate data from the next 10 subjects, corrects the annotations and retrains the hHMM with the 30 subjects, and so on.
& Reduces the annotation time to only 15\%.
& Requires 15\% of labeled data. \\ \hline

\makecell{\cite{ponnada2019designing}\\ (2019)}
& Real-time activity annotation with a custom tablet app.
& Decreases the annotation time.
& Baseline requires pre-labeled data.\\ \hline

\makecell{\cite{faridee2019augtoact}\\ (2019)}
&Semi-Supervised/TL module + Augmentation model.
&Requires a low amount of labeled data.
&Requires labeled data and works only for previously seen activities.\\ \hline

\makecell{\cite{hossain2019active}\\ (2019)}
& Actor-critic network that annotates data based on partially manually annotated data, starting from videos and data annotated by users of a mobile app. 
& Maintains the same results obtained with 100\% labeled data using less than 30\% labeled data.
& Requires user intervention and labeled data.\\ \hline

\makecell{\cite{sawano2020annotation}\\ (2020)}
& Smartphone app that collects data and annotates the users' response to notifications.
& Reduces the annotation time.
& Requires user feedback and labeled data.\\ \hline

\makecell{\cite{kwon2020imutube}\\ (2020)}
& Utilizes existing large-scale video repositories to generate virtual IMU data for training the HAR system.
& Generates IMU data from video.
& Requires a training set comprising of video and IMU data.\\ \hline

\makecell{\cite{avsar2021benchmarking}\\ (2021)}
& Semiautomated annotation combining predictions from CNN-IMU with human revision.
& Reduces the annotation time.
& Requires human revision.\\ \hline

\makecell{\cite{korpela2021reducing}\\ (2021)}
& Semi-automatic labeling process.
& Reduces the annotation time.
& Requires human revision.\\ \hline

\makecell{\cite{tang2021selfhar}\\ (2021)}
& Adopts a teacher-student setup where a teacher model distills knowledge from labeled data by annotating a large unlabeled dataset.
& Reduces the annotation time and the required data by a factor of 10, and was tested on 9 different HAR datasets.
& Requires human revision.\\ \hline

\makecell{\cite{yhdego2022fall}\\ (2022)}
& Uses VR falls to capture real falls.
& Provides realistic falls.
& Uses synthetically generated labels.\\ \hline

\makecell{\cite{mohamed2022har}\\ (2022)}
& HAR-GCNN is a graph comprised of partially labeled sensor measurements that represent chronologically ordered activities.
& Improves the classification accuracy by about 25\% and up to 68\% on different datasets.
& Requires user intervention.\\\hline \hline
\end{tabular}}
\end{table*}

In their study, Saeedi et al.~\cite{saeedi2017co} propose a multi-expert mobile health system utilizing AL techniques. The architecture addresses challenges related to reconfiguring mobile sensor devices for health monitoring. One challenge is the expensive cost of data labeling, which can interfere with the user's life and requires feedback from healthcare experts or costly equipment. Another challenge involves identifying the most suitable expert for each data instance, and a third challenge is the uncertainty of labels due to limited expert knowledge.
To overcome these challenges, the proposed architecture selects the most cost-effective and confident expert for each query, considering collaboration among experts to minimize cost and improve data labeling accuracy. The authors also develop new algorithms for system initialization, utilizing clustering algorithms and ensemble classification methods.
The effectiveness of the architecture and algorithms is demonstrated through a case study on activity monitoring, achieving a 85\% accuracy in activity recognition by labeling only 15\% of unlabeled data and reducing annotation costs.
Future work aims to enhance the architecture to handle multi-modality sensory systems, to conduct a real pilot study, and to integrate TL and AL approaches to further reduce the number of queries and improve the learner accuracy.


In \cite{martindale2018smart}, Martindale et al. proposed a HAR pipeline for semi-automated labeling and efficient collection of daily activity data. This data was labeled by identifying the walking cycle phases based on  video, IMU, and pressure insole data. Although the setup was designed in a controlled environment, the same principles could be applied to more natural or specific applications. The authors proposed an on-the-edge video detection method to detect on-the-ground and off-the-ground stride phases with only 17\% manual labeling or correction required. 
This technique reduced the labeling time by 83\% compared to complete manual labeling without assistance. 

Gan et al., in \cite{gan2018automatic}, devised two semi-automated labeling algorithms: one for personalized fall detection online model training using k-Means clustering, and another one for personalized localization online model training, which employed a descendingly ordered set of peak-trough magnitudes. Based on their experiments, the authors reported that the proposed approach resulted in a fall detection accuracy of up to 97\%. In contrast, the labeling of peak-trough magnitudes led to a step counting accuracy of at least 96\%.

Bota et al., in \cite{bota2019semi}, proposed a Semi-Supervised AL (SSAL) approach to address the challenges posed by the significant volume of data recorded by unobtrusive and pervasive sensors, such as smartphones and wearables. The proposed approach consists of two steps: (1) selecting the most relevant samples to be labeled by an expert using a Query Strategies (QSs) criterion, and (2) propagating the  labels of annotated samples to similar samples on the entire dataset using an automatic method. 
The study was tested on two HAR datasets using a comprehensive study of state-of-the-art QS and Stopping Criteria (SC) techniques, and a comparison to AL. 
The methods were evaluated over several automatic annotation strategies based on different distance functions to optimize the SSAL model. 
This paper extended the work conducted by \cite{stikic2009activity} on HAR by applying Self-Training (ST) on the labels previously selected by AL. 

In \cite{martindale2019hidden}, Martindale et al. proposed a pipeline to overcome the lack of realistic and labeled datasets for medical applications of cyclic activity monitoring, such as step-counting and gait analysis. 
The pipeline reduces the percentage of labels that require manual adjustment to only 14\%, making it possible to produce a public dataset of over 150,000 labeled cycles from 80 participants. The dataset includes 12 activities, 10 of which are cyclic, and features diverse ranges of bouts, transitions, and non-straight walking. For datasets related to e.g., home monitoring, where mostly walking data is expected, the labeling effort for new datasets can be as low as 8\%. 
Furthermore, the authors proposed an iterative training technique for a hierarchical Hidden Markov Model (hHMM) in this paper. The hHMM hierarchy includes cycle phases for each of the 10 cyclic activities, and this method allowed the dataset, which has been made publicly available in \cite{martindale2018smart}, to be expanded fourfold with a final miss rate of 0.6\% and a false discovery rate of 7.6\%. The complete pipeline achieved an F1-score of 89.5\%, with an expected F1-score for new data of 93.0\%. 

In \cite{ponnada2019designing}, Ponnada et al. introduced two design prototypes for Human Computation Games (HCG), namely Mobots and Signaligner, with the purpose of motivating players to label raw accelerometer data. These games were trained using annotated data, which provided players with an initial assessment of the accelerometer data provided. The objective for Mobots players was to annotate data fragments with activity names, while Signaligner players aimed to match input data patterns with visual pattern templates. In terms of performance, Mobots players successfully annotated 8.7 hours of accelerometer data using only 9.5 minutes of annotated data, achieving an overall accuracy of 89.7\%. On the other hand, Signaligner players achieved a 99.5\% accuracy in labeling 11.69 hours of acceleration data, starting from 3.8 hours of annotated data.
This difference in performance was attributed to the fact that Signaligner players were provided with more signal context and visual patterns to match with an on-screen reference, whereas Mobots players had to rely on their memory of signals and activity categories to label short data fragments.

Faridee et al.\cite{faridee2019augtoact}  introduced the AugToAct framework, a flexible and innovative semi-supervised TL framework with augmented capabilities. 
The framework can be applied to different classification and domain adaptation tasks, showcasing its suitability for complex HAR. In particular, the proposed technique, starting from an annotated dataset, aims to augment the dataset with artificial data samples labeled as the initial dataset. 
The authors aim to automate the process of identifying optimal augmentation parameters in future research. 
Additionally, they plan to evaluate the model's generalizability on a broader range of datasets, encompassing diverse human activities. 
Notably, the current experiment does not address unseen labels in the target domain, a limitation the authors plan to overcome in their future work.

In \cite{hossain2019active}, Hossain et al., proposed a DL model for activity recognition that incorporates AL in hyperparameter tuning, as opposed to previous works that only focused on identifying the most informative instance through AL. To achieve this, the authors suggested optimizing network parameters using a joint loss function that combined the cross-entropy loss of the DL model and the entropy function of the AL pipeline. To validate their approach, they used a mobile application to collect data in real-world settings, and the results showed that the joint loss function helped the DL model to generalize better with lower weight values, even in the presence of outliers. 
The authors also introduced an annotator selection model based on the contextual similarity between annotators and users, outperforming other algorithms by converging faster into optimal accuracy. 

Sawano et al. \cite{sawano2020annotation} proposed a method for estimating the user and device status, based on user responses to notifications generated by a smartphone. The experiments showed that the proposed method had an average precision of 76.9\% and 96.3\% for user-independent and user-dependent experiments, respectively. Although the proposed method had a high annotation precision, the recall was low, meaning that accurate annotations can only be assigned to a limited amount of data. However, since the method has an automatic annotation collection mechanism, it can collect a large amount of annotated data for many people over a long period of time.

In \cite{kwon2020imutube}, Kwon et al. addressed the lack of labeled data in HAR by introducing IMUTube. This automated processing pipeline generates virtual streams of IMU data from human activity videos.  
The authors demonstrated the effectiveness of the virtually-generated IMU data in improving the performance of existing HAR models. 

Avsar et al. \cite{avsar2021benchmarking} present an approach for generating high-quality data in the context of multi-channel time series HAR. Their method utilizes optical motion capturing and inertial measurements from on-body devices to combine temporal CNN predictions with manual revisions, resulting in fine-grained annotations.
The approach was evaluated in terms of time consumption and annotation consistency, revealing a substantial reduction in annotation effort by up to 62.8\%.

Korpela et al., in \cite{korpela2021reducing}, propose a technique that utilizes an image segmentation algorithm called SLIC (Simple Linear Iterative Clustering) to perform temporal clustering of the classifier output. 
The time-series data was fed to the algorithm as a 1D image, with the class probabilities serving as the color channels. The proposed method was evaluated on 233 minutes of time-series data, and it achieved an average reduction of 56\% in annotation time compared to the baseline method that used raw classifier output. 

In \cite{tang2021selfhar}, Tang et al., proposed SelfHAR, a semi-supervised model that leverages unlabeled mobile sensing datasets to improve the performance of HAR models. SelfHAR uses a combination of teacher-student self-training and multi-task self-supervision to learn robust signal-level representations and augment small labeled datasets. 
This technique was evaluated on various HAR datasets and outperformed other supervised and semi-supervised approaches, achieving up to a 12\% increase in F1-score with the same number of model parameters at inference. 
Additionally, SelfHAR achieved similar performance by using up to 10 times less labeled data than supervised approaches. 

In their work on fall detection \cite{yhdego2022fall}, Yhdego et al. introduced a self-supervised learning approach that utilizes unlabeled data to pre-train Fully Connected Network (FCN) and Residual Neural Network (ResNet) models. These pre-trained models are then fine-tuned using labeled data. The method incorporates overlapping sliding windows for feature extraction and addresses the issue of imbalanced classes in the dataset through oversampling and a modified weighted focal loss function.
Experimental results demonstrated that the ResNet self-supervised DL method, combined with random oversampling, achieved an impressive average F1-score of 98\% for accurately detecting falls.

Mohamed et al., in \cite{mohamed2022har}, present HAR-GCCN (HAR-Graph Chronologically Correlation Network), a deep graph CNN model for HAR using mobile sensor data. 
They proposed leveraging the implicit chronology of human behavior to learn unknown labels and classify future activities. This was done using a new training strategy that predicts missing activity labels by leveraging the known ones. 
HAR-GCCN outperformed baseline methods, improving classification accuracy by up to 68\% on different datasets. In addition, they reported that HAR-GCNN has stable performance, independently of the number of chronologically ordered activities considered within the input graph.

\subsubsection{Environment-driven approaches}\label{sec:4.1.2}
We next describe an approach to semi-automatic annotation techniques that makes primarily use of human and environment-driven knowledge to annotate HAR data. 
Tables \ref{tab:S-ED} and \ref{tab:S-ED-1} offer a comprehensive summary of the 4 articles falling into such category. 
One of the first environmental-based, semi-automated methodologies has been published by Szewcyzk et al. \cite{szewcyzk2009annotating}\footnote{In this article, the authors explore an annotation technique falling both into semi- and fully- automated environment-driven categories.}. 
Szewcyzk et al. explored four alternative mechanisms for annotating sensor data with corresponding activity labels to monitor the functional health of smart home residents. The first method utilizes the raw data from sensors along with a map of the apartment to identify the activities being performed. The location of the sensors and the time of day are used to infer the activities. For example, motion and water sensors triggered during a specific time could indicate meal preparation. 
In the second method, the residents provide time diaries reporting their activities every half an hour. This approach is less invasive than others but relies on the residents' self-reports, which may not always be reliable.
The third and fourth methods involve using a visualization tool to analyze the sensor events. Method 3 uses the visualization tool for manual annotation, while method 4 includes resident feedback. A 3D environment simulator called CASASim displays sensor readings in real-time. Researchers rely on the combined information from the simulator and resident time diaries to interpret and annotate the sensor events.

Subramanya et al.~\cite{subramanya2012recognizing} proposed a dynamical graph model to jointly estimate activity and spatial context over time, based on asynchronous observations from GPS measurements and a wearable sensor. 
The graph model's parameters are trained on partially labeled data, and the authors applied virtual evidence to improve data annotation, providing high flexibility in labeling training data. 
Experiments suggest that the proposed system achieves a recognition accuracy of 95\%. This is significantly higher than existing techniques that do not perform joint reasoning about a person's activities and spatial context. 
\begin{table*}[!t]
\centering
\caption{Semi-automated environment-driven approaches.}\label{tab:S-ED}
\resizebox{\textwidth}{!}{
\begin{tabular}{c||cccccccccc}
\hline
\makecell{\textbf{Ref.}\\ \textbf{(Year)}} & \makecell{\textbf{Dataset}\\\textbf{Provided (\#)}} & \textbf{Devices} & \textbf{Sensors} & \makecell{\textbf{\# of}\\ \textbf{Sensors}} & \textbf{Position} & \textbf{Model} & \makecell{\textbf{\# of} \\\textbf{Subjects [M/F]}} & \textbf{Activities} & \makecell{\textbf{\# of} \\\textbf{Activities}} & \makecell{\textbf{In-Out}\\ \textbf{Door}}\\ \hline \hline

\makecell{\cite{szewcyzk2009annotating}\footref{fo:54}\\ (2009)}
&yes (1)
&environment
&\makecell{motion sensors, \\temperature}
&27
&\makecell{phone book, cooking pot, \\medicine container, \\cooking ingredients pot}
&\textit{n/a}
&2[2/0]
&OBI-ADL
&6
&Indoor\\ \hline

\makecell{\cite{subramanya2012recognizing}\\ (2012)}
&yes (1)
&standalone
&\makecell{GPS, accelerometer,\\ 2 microphones, brightness,\\ temperature, barometer} 
&7
&\makecell{left\\ shoulder}
&\makecell{graph\\ model} 
&6
&\makecell{BHA, OBI}
&6
&\makecell{Indoor+\\ Outdoor}
\\ \hline

\makecell{\cite{woznowski2017talk}\\ (2017)}
&yes (1)
&smartphone
&\makecell{mobile app, voice,\\ location, NFC}
&1
&\textit{n/a}
&\textit{n/a}
&\textit{n/a}
&\makecell{ADL}
&\textit{n/a}
&Indoor
\\ \hline

\makecell{\cite{tonkin2018talk}\\ (2017)}
&yes (1)
&smartphone
&\makecell{mobile app, voice,\\ location, NFC}
&1
&\textit{n/a}
&\textit{n/a}
&10
&\makecell{ADL}
&29
&Indoor
\\ \hline

\makecell{\cite{solis2019human}\\ (2019)}
&yes (1)
&smartwatch
&Bluetooth antenna
&1
&wrist
&\makecell{graph-based BLE-\\location mapping}
&12
&\makecell{BHA}
&1
&Indoor
\\ \hline \hline
\multicolumn{11}{c}{In Columns Devices, \# of Sensors, Position, \# of Subjects [M/F], Activities, and \# of Activities the symbol \"-\" is used to separate information of the different datasets.}
\end{tabular}}
\end{table*}

\begin{table*}[!t]
\centering
\caption{Annotation method, Advantage's, and Limitations of Semi-automated environment-driven approaches.}\label{tab:S-ED-1}
\resizebox{\textwidth}{!}{
\begin{tabular}{c||p{7cm}p{5cm}p{5cm}}
\hline
\textbf{Ref. (Year)} & \textbf{Annotation} & \textbf{Advantages} & \textbf{Limitations} \\ \hline \hline

\makecell{\cite{szewcyzk2009annotating}\footref{fo:54}\\ (2009)}
& Four distinct annotation/visualization systems were employed, each with its own approach: raw data and hard-coded recognition, raw data and resident time diaries, visualization of the sensor data tool, and visualization of the sensor data tool combined with residents' feedback.
& Explored the topic and conducted a preliminary comparison of various annotation systems.
& Recogniton not performed and requires human labels and intervention\\ \hline

\makecell{\cite{subramanya2012recognizing}\\ (2012)}
& Utilizes a dynamic graphical model to jointly estimate activity and spatial context over time.
& Achieves 95\% accuracy in recognizing activities by leveraging environmental information.
& Requires labeled data.\\ \hline

\makecell{\cite{woznowski2017talk}\\ (2017)}
& Mobile app that reads NFC tags attached to daily objects and places for almost automatic activity annotation.
& Reduces effort for data annotation.
& Requires user feedback.\\ \hline

\makecell{\cite{tonkin2018talk}\\ (2017)}
& Mobile app that reads NFC tags attached to daily objects and places for almost automatic activity annotation.
& Reduces effort for data annotation.
& Requires user feedback.\\ \hline

\makecell{\cite{solis2019human}\\ (2019)}
& Uses Bluetooth devices to recognize specific environmental positions and asks the user on their smartwatch if they are performing a specific activity.
& Reduces effort for data annotation.
& Requires user feedback.\\ \hline \hline
\end{tabular}}
\end{table*}

In the studies conducted by Woznowski et al. \cite{woznowski2017talk} and Tonkin et al. \cite{tonkin2018talk}, various approaches were explored to allow users to self-annotate their activities in near-real-time for the development of accurate HAR algorithms. 
The study proposed a mobile app with multiple logging capabilities for self-annotation of activities. These capabilities included model-based, voice-based, location-based, and NFC-based methods. Users interacted directly with the app, except for the NFC-based approach, which was fully automatic upon contact with NFC tags.


Finally, in \cite{solis2019human}, Solis et al. proposed a methodology to improve the recognition of activities related to eating using wearable computers in natural environments. They utilized location information from wearable sensors in IoT platforms to learn the users' behavior patterns without prior knowledge. Annotations were requested only when automatic annotation failed. In a case study with 12 participants wearing smartwatches, audio recordings were used for labeling eating moments. The study showed a 2.4\% accuracy improvement with a limit of 20 requested annotations per day. 
A dietary monitoring study validated the algorithm based on classifier uncertainty, allowing long-term data collection with minimal annotations.

\subsubsection{Hybrid}\label{sec:4.1.3}
The focus of hybrid methodologies is to further reduce the manual effort and expenses associated with the annotation process by combining data and environmental information, while enhancing the accuracy and performance of activity recognition models. 
Tables \ref{tab:S-H} and \ref{tab:S-H-1} provide an overview over the 6 methodologies falling into this category.

With this aim, in \cite{alam2015mobeacon}, Alma et al. proposed Mobeacon: a mobile phone and iBeacon sensor-based smart home activity recognition system, which uses Bagging Ensemble Learning (BEL) and Packaged Naive Bayes (PNB) classification algorithms for high-level activity recognition on smartphones. 
The authors incorporated the semantic knowledge of the testing environment and use it with the built-in adaptive learning models on the smartphone to facilitate the ground truth data annotation. They demonstrated that Mobeacon outperforms existing lightweight activity recognition techniques in terms of accuracy (max. 94\%) in a low-resource scenario and is sufficiently efficient to reside on smartphones for recognizing ADLs in real-time. The authors also designed an efficient smartphone application interface for defining and creating an initial semantic knowledge base about the smart home environment. They used their semantic knowledge base, expression tree-based activity construction, and an inference cache to accelerate the activity recognition process of their lightweight BEL-based approach. 

Meurisch et al. \cite{meurisch2015labels} proposed "Labels," a self-tracking mobile application that provides a user interface for annotating automatically collected sensor data from mobile, desktop, and social media platforms with metadata, such as performed activities. The study evaluated Labels with 163 participants over a four-week field study, collecting over 43,000 manually annotated data samples. Results show that the participants annotated about 82.5\% of their place-related time slots with their performed activities. 

Nino et al.~\cite{nino2016scalable} presented a methodology for the semi-automatic generation of reliable position annotations to evaluate multi-camera people trackers on large video data sets. The methodology automatically computed most of the annotation data by recognizing the person's position and interaction with daily life objects. 
The proposed framework is generic and can handle additional trackers. 
The authors provided guidelines on applying the proposed methodology to new data sets and presented an exploratory study for the multi-target case. 

Cruciani et al.~\cite{cruciani2018personalized} proposed an annotation system that integrates GPS and a step counter as two information sources. The GPS data is utilized to differentiate activities based on position, estimated speed, and predefined heuristics. Speed ranges associated with each activity (e.g., walking: 1.4 - 2.0 m/s, running: 3.0 - 6.0 m/s, transportation: > 8 m/s) are employed for labeling activities. To enhance the accuracy and reduce mislabeled samples, a step counter is incorporated into the system.
The system combines the GPS and step counter data through a rule-based intersection of these information sources. This combination allows for more precise labeling and discrimination between activities. For instance, it can distinguish between running and driving a vehicle, or detect running activities in a gym environment that may not be identifiable using GPS alone.

\begin{table*}[!t]
\centering
\caption{Semi-automated hybrid approaches.}\label{tab:S-H}
\resizebox{\textwidth}{!}{
\begin{tabular}{c||cccccccccc}
\hline
\makecell{\textbf{Ref.}\\ \textbf{(Year)}} & \makecell{\textbf{Dataset}\\\textbf{Provided (\#)}} & \textbf{Devices} & \textbf{Sensors} & \makecell{\textbf{\# of}\\ \textbf{Sensors}} & \textbf{Position} & \textbf{Model} & \makecell{\textbf{\# of} \\\textbf{Subjects [M/F]}} & \textbf{Activities} & \makecell{\textbf{\# of} \\\textbf{Activities}} & \makecell{\textbf{In-Out}\\ \textbf{Door}}\\ \hline \hline
\makecell{\cite{alam2015mobeacon}\\ (2015)}
&yes (1)
&\makecell{smartphone,\\ BLE beacons,\\ video}
&\makecell{accelerometer,\\ gyroscope,\\ RSSI}
&13
&\makecell{smartphone in front\\ right pocket, beacons on\\ objects and positions \\of a home environment}
&\makecell{semantic\\knowledge}
&2 [1/1]
&\makecell{BHA, OBI, ADL}
&25
&indoor
\\ \hline

\makecell{\cite{meurisch2015labels}\\ (2015)}
&yes (1)
&smartphone
&\makecell{accelerometer, location, light,\\ proximity, gyroscope, loudness,\\ wifi signals, smartphone usage, charging\\ state, social network information,\\ user-annotated human activity data}
&1
&\textit{n/a}
&\textit{n/a}
&163
&BHA
&\textit{n/a}
&\makecell{indoor+\\outdoor}\\ \hline

\makecell{\cite{nino2016scalable}\\ (2016)}
&yes (1)
&video
&video
&4
&\textit{n/a}
&\makecell{poselet-based\\ people detector}
&1
&\makecell{BHA, OBI}
&8
&indoor\\ \hline

\makecell{\cite{cruciani2018personalized}\\ (2018)}
&yes (1)
&smartphone
&accelerometer, step counter, GPS
&1
&trouser pocket
&\textit{n/a}
&1
&\makecell{BHA}
&4
&\makecell{indoor+\\outdoor}\\ \hline

\makecell{\cite{cruz2019semi}\\ (2019)}
&yes (1)
&smartwatch
&\makecell{accelerometer,\\ gyroscope,\\ audio}
&2
&wrist
&\makecell{DTW,\\ SVM,  SMO}
&15
&\makecell{hand gestures,\\ - recorded audio answers}
&36-20
&indoor\\ \hline

\makecell{\cite{tonkin2019towards}\\ (2019)}
&yes (1)
&\makecell{smartwatch,\\ environment}
&\makecell{temperature, humidity,\\ water usage, electricity\\ usage per plug, and\\ PIR sensor, accelerometer}
&\textit{n/a}
&enviroment and wrist
&\makecell{natural language\\toolkits} 
&1
&\makecell{ADL, OBI}
&9
&indoor\\ \hline \hline 
\multicolumn{11}{c}{In Columns Devices, \# of Sensors, Position, \# of Subjects [M/F], Activities, and \# of Activities the symbol \"-\" is used to separate information of the different datasets.}
\end{tabular}}
\end{table*}

\begin{table*}[!th]
\centering
\caption{Annotation method, advantages, and limitations of semi-automated hybrid approaches.}\label{tab:S-H-1}
\resizebox{\textwidth}{!}{
\begin{tabular}{c||p{7cm}p{5cm}p{5cm}}
\hline
\textbf{Ref. (Year)} & \textbf{Annotation} & \textbf{Advantages} & \textbf{Limitations} \\ \hline \hline
\makecell{\cite{alam2015mobeacon}\\ (2015)}
& Annotates sensor data based on recognized location and used objects.
& Reduces labeling effort.
& Requires preliminary knowledge of activity semantics and user feedback.\\ \hline

\makecell{\cite{meurisch2015labels}\\ (2015)}
& Manual labeling through a smartphone app with some automation:\linebreak
1) Pre-segmentation of data based on location change, requiring manual labeling of each segment.
2) Quality metrics for labeling with additional custom tags.
& Provides a dataset.
& Requires user interaction with the smartphone.\\ \hline

\makecell{\cite{nino2016scalable}\\ (2016)}
& Recognizes user positions and interaction with objects in the environment.
& Automatically annotates 80\% of the video frames with 99\% accuracy.
& Works only with one person in the environment. Relies on a pre-trained model for object recognition and distance measurement.\\ \hline

\makecell{\cite{cruciani2018personalized}\\ (2018)}
& Generates weak labels through a heuristic combining accelerometer and GPS data.
& Achieves an overall accuracy of 87\% compared to 74\% with fully supervised approaches.
& Requires user interaction with the smartphone.\\ \hline

\makecell{\cite{cruz2019semi}\\ (2019)}
& Uses a smartwatch to recognize hand gestures and the position for annotating complex activities, and a microphone to recognize specific home contexts and to annotate data from wearables.
& Reduces labeling effort.
& Requires preliminary annotated data for hand gesture scenarios.\\ \hline

\makecell{\cite{tonkin2019towards}\\ (2019)}
& Audio annotation of the start time of an activity.
& Easier than manual video-based annotation.
& Requires user interaction with the system.\\ \hline \hline
\end{tabular}}
\end{table*}

In~\cite{cruz2019semi}, the authors proposed two approaches for semi-automated online data labeling. The first approach is based on the recognition of subtle finger gestures performed in response to a data-labeling query. In contrast, the second approach focuses on labeling activities with an auditory manifestation and uses a classifier to estimate the activity and a conversational agent to ask the participant for clarification or additional data. Results show that while both studies have limitations, they achieve a precision from 80\% to 90\%. In addition, the authors described an approach for the semi-automatic labeling of environmental audio data and presented the results of experiments to assess its feasibility. 

In~\cite{tonkin2019towards}, the authors proposed a participant-centric free-text annotation process to facilitate activity recognition in a kitchen environment and characterized the resulting annotations. They reviewed the data from the study for assessing the complexity of  cooking activities using the dataset, and found that the annotations explored in the paper constitute a useful basis for an exploratory analysis of the data. 
However, they noted that the granularity of the annotations is not optimal for certain tasks and may benefit from a more detailed set of annotations. The authors also identified several features of meal preparation complexity that are readily detectable in their sensor data, including monitored appliance use, water use, and the energy released as heat and humidity during the task. 

\subsection{Fully automated}\label{sec:4.2}
This section will discuss the 14 papers on fully-automated data annotation techniques for HAR in the three previously mentioned categories and provide detailed descriptions of their proposed methods and key features. As shown in Table \ref{tab:all-papers}, none of these papers focus on hybrid fully-automated methodologies. Therefore, this section will focus solely on data-driven and environment-driven methods. 

\subsubsection{Data-driven approaches}\label{sec:4.2.1}
These studies propose fully-automated data-driven methodologies for automated data annotation in HAR, leveraging the existence of data patterns being extracted through various techniques such as AL, augmented and TL frameworks, or self-supervised learning.
Tables \ref{tab:H-DD} and \ref{tab:H-DD-1} provide an overview of the 9 publications falling into this category.

\begin{table*}[!th]
\centering
\caption{Fully-automated data-driven based approaches.}\label{tab:H-DD}
\resizebox{\textwidth}{!}{
\begin{tabular}{c||c
ccccccccc}
\hline
\makecell{\textbf{Ref.}\\ \textbf{(Year)}} & \makecell{\textbf{Dataset}\\\textbf{Provided (\#)}} & \textbf{Devices} & \textbf{Sensors} & \makecell{\textbf{\# of}\\ \textbf{Sensors}} & \textbf{Position} & \textbf{Model} & \makecell{\textbf{\# of} \\\textbf{Subjects [M/F]}} & \textbf{Activities} & \makecell{\textbf{\# of} \\\textbf{Activities}} & \makecell{\textbf{In-Out}\\ \textbf{Door}}\\ \hline \hline
\makecell{\cite{jardim2016automatic}\\ (2016)}
&yes (1)
&Kinect
&\makecell{depth\\ image}
&1
&body joints
&\makecell{segmentation\\ and temporal\\ clustering}
&12
&\makecell{PA}
&8
&indoor\\ \hline

\makecell{\cite{rokni2018autonomous}\\ (2018)}
&yes (1)
&\makecell{Xsens\\ MTx}
&\makecell{accelerometer,\\ gyroscope,\\ magnetometer}
&5
&\makecell{torso, right arm,\\ left arm, right leg,\\ left leg}
&TL
&8 [4/4]
&\makecell{BHA}
&15
&\makecell{indoor +\\outdoor}\\ \hline

\makecell{\cite{liang2018automatic}\\ (2018)}
&yes (1)
&IMU
&\makecell{accelerometer,\\ gyroscope}
&9
&\makecell{left upper arm,\\ right upper arm,\\ left wrist, right wrist,\\ front waist, left thigh,\\ right thigh, left ankle,\\ and right ankle}
&\textit{n/a}
&6 [5/1]
&\makecell{BHA, R}
&17
&indoor\\ \hline

\makecell{\cite{zhang2020deep}\\ (2020)}
&no (3)
&video, IMU
&\makecell{video,\\ accelerometer,\\ gyroscope}
&5
&\makecell{arms, legs\\ and back}
&\makecell{cVAE\\ cGAN}
&100
&\makecell{ADL}
&\textit{n/a}
&\makecell{indoor +\\outdoor}\\ \hline

\makecell{\cite{jeng2021wrist}\\ (2021)}
&yes (1)
&IMU
&accelerometer
&2
&\makecell{wrist,\\ chest}
&\makecell{knowledge-\\ based} 
&2
&\makecell{sleeping\\ posture}
&4
&indoor\\ \hline

\makecell{\cite{ullrich2021detection}\\ (2021)}
&yes (1)
&IMU
&\makecell{accelerometer,\\ gyroscope}
&2
&insoles
&\makecell{lnowledge-\\ based} 
&12 [10/2]
&\makecell{gait\\ phases}
&\textit{n/a}
&\makecell{indoor +\\outdoor}\\ \hline

\makecell{\cite{ma2021unsupervised}\\ (2021)}
&no (3)
&\makecell{smartwatch,\\ smartphone\\ - smartphone\\ - smartphone}
&\makecell{accelerometer,\\ gyroscope}
&\makecell{2 - 1\\ - 1}
&\makecell{pocket, wrist\\ - pocket\\ - pocket}
&\makecell{CNN-BiLSTM\\ autoencoder,\\ k-means}
&36 - 66 - 24
&\makecell{BHA, ADL, R}
&6 - 15 - 6
&indoor \\ \hline

\makecell{\cite{qi2022dcnn}\\ (2022)}
&yes (1)
&\makecell{smartphone,\\ kinect}
&\makecell{accelerometer,\\ gyroscope,\\ magnetometer,\\ depth image}
&2
&\makecell{arm, 25\\ body joints}
&\makecell{hierarchical\\ k-medoids} 
&10 [6/4]
&\makecell{BHA, PA}
&12
&indoor\\ \hline

\makecell{\cite{lin2022clustering}\\ (2022)}
&no (1)
&smartphone
&\makecell{accelerometer,\\ gyroscope}
&1
&waist
&FS-FCPSO
&30
&\makecell{BHA}
&6
&indoor\\ \hline \hline
\multicolumn{11}{c}{In Columns Devices, \# of Sensors, Position, \# of Subjects [M/F], Activities, and \# of Activities the symbol \"-\" is used to separate information of the different datasets.}
\end{tabular}}
\end{table*}

\begin{table*}[!th]
\centering
\caption{Annotation method, Advantages, and Limitations of Fully-automated data-driven based approaches.}\label{tab:H-DD-1}
\resizebox{\textwidth}{!}{
\begin{tabular}{c||
p{7cm}
p{5cm}
p{5cm}}
\hline
\textbf{Ref. (Year)} & \textbf{Annotation} & \textbf{Advantages} & \textbf{Limitations} \\ \hline \hline
\makecell{\cite{jardim2016automatic}\\ (2016)}
& Extracts features related to body joints and utilizes knowledge of performed activities to recognize and label them automatically.
& Automatically generates synthetic data from video.
& Limited set of recognized activities. Difficulties in scaling this to other activities. \\ \hline

\makecell{\cite{rokni2018autonomous}\\ (2018)}
& Utilizes synchronous multi-view Learning, including labeling target instances by semi-labels MEEL, clustering target instances, building a weighted bipartite graph, and propagating cluster labels to their instances.
& Does not require user feedback and achieves a recognition accuracy of 84\% on unlabeled data.
& Lower accuracy compared to labeled data. \\ \hline

\makecell{\cite{liang2018automatic}\\ (2018)}
& Automatically labels activities based on mathematical transformations for detecting activity endpoints, performed in three steps: preliminary segmentation, endpoint detection, and label assignment.
& Does not require user feedback and achieves a recognition accuracy of 89\% on unlabeled data.
& Limited set of recognized activities. Difficulty in scaling to other activities. \\ \hline

\makecell{\cite{zhang2020deep}\\ (2020)}
& Uses a dataset containing video and sensor data to train the model and generate synthetic data for the studied activities.
& Generates synthetic motion data of specific activities from videos.
& No system accuracy is provided. \\ \hline

\makecell{\cite{jeng2021wrist}\\ (2021)}
& Utilizes a chest-mounted accelerometer to annotate sleeping and non-sleeping postures.
& Data annotation is fully automated.
& Specific to sleeping activities. \\ \hline

\makecell{\cite{ullrich2021detection}\\ (2021)}
& Knowledge-based annotation.
& Achieves a 94\% F1-score in recognition.
& Knowledge-based. \\ \hline

\makecell{\cite{ma2021unsupervised}\\ (2021)}
& Utilizes unsupervised learning through multi-task deep clustering.
& Achieves an average F1-Score higher than 85\%.
& Less performant than supervised techniques. \\ \hline

\makecell{\cite{qi2022dcnn}\\ (2022)}
& Utilizes unsupervised learning through Hk-Medoids.
& Achieves an accuracy of 94\% in labeling.
& - \\ \hline

\makecell{\cite{lin2022clustering}\\ (2022)}
& Feature selection technique based on FS-FCPSO.
& Performs better than the classic k-Means algorithm.
& Less performant than supervised techniques. \\ 
\hline \hline
\end{tabular}}
\end{table*}

The initial study that introduced fully-automatic data-driven annotation techniques for HAR was conducted by Jardim et al. in \cite{jardim2016automatic}. The researchers proposed a method for recognizing human actions from a continuous sequence of images captured by a Kinect sensor. They designed an automatic temporal segmentation approach to divide the sequence into individual actions and employed a straightforward filtering technique based on joint movement. Furthermore, they presented an automatic labeling method utilizing a clustering algorithm on a subset of available features. 
To enhance the outcomes, they recommended the utilization of Euler angles and dynamic time warping (DTW) techniques. They successfully demonstrated that combining clustering and filtering techniques allows for the unsupervised labeling of human actions captured by a depth-sensing camera that tracks skeleton body joints.

In~\cite{rokni2018autonomous}, Rokni et al. introduced an autonomous multi-view learning approach capable of dynamically retraining ML algorithms in real-time without the need for labeled training data. 
By employing the approach in batch mode, they achieved an 83.7\% accuracy in activity recognition, representing a 9.3\% improvement facilitated by automatic data labeling in the new sensor node. Instead, in the online mode, it achieved an 82.2\% accuracy in activity recognition. 
This study represents an initial step towards developing next-generation wearables with computational autonomy and automatically learning ML algorithms. 

A paper by Liang et al.~\cite{liang2018automatic} presents ALF, an Automatic Labeling Framework for in-laboratory HAR, eliminating the need for a small initial set of labeled data. The proposed framework converts time series activity data into absolute wavelet energy entropy and detects activity endpoints using constraints and information extracted from a predefined human activity sequence. 
The authors evaluated the framework's performance on a collected dataset and the UCI HAR dataset\cite{anguita2013public}, achieving average precision and recall scores above 81.9\%, and average F-measure scores above 88.9\%. The ALF framework significantly reduces labeling efforts, while maintaining the labeling accuracy. It provides a fast and reliable method for generating labeled datasets, with a total labeling time of approximately 18.6 minutes, which is 75.8\% shorter than the average manual labeling time of 76.8 minutes.

In~\cite{zhang2020deep}, Zhang et al. proposed two deep generative cross-modal architectures to synthesize accelerometer data streams from video data streams. The approach utilizes a conditional generative adversarial network (cGAN) to generate sensor data based on video data and incorporates a conditional variational autoencoder (cVAE)-cGAN to further enhance the data representation. 
The proposed method was evaluated through experiments on publicly available sensor-based activity recognition datasets, comparing models trained on synthetic data against those trained on real sensor data. 

In~\cite{jeng2021wrist}, Jeng et al. introduced iSleePost, a sleep posture monitoring system for home care that automatically recognizes body posture during sleeping for labeling data. By analyzing data from a single wrist sensor, the system achieves an accuracy of up to 85\% in posture recognition. 
The authors evaluated two different learning algorithms, with the RF algorithm achieving over 70\% accuracy and the SVM algorithm achieving 73\% accuracy. iSleePost is more cost-effective than existing approaches relying on pressure mats, cameras, or specialized equipment. 

In~\cite{ullrich2021detection}, the authors introduced a pipeline for the automated detection of unsupervised standardized gait tests from continuous real-world IMU data. 
The proposed approach involves gait sequence detection, peak enhancement, and subsequence DTW to identify gait test series, which are further decomposed into individual 4 x 10 meters walking-tests. These tests were used to assess the walking velocity. The algorithm was evaluated using 419 gait test series, achieving an F1-score of 88.9\% for detection and 94.0\% for decomposition. 

In~\cite{ma2021unsupervised}, Ma et al., presented an end-to-end multi-task deep clustering framework that integrates feature representation, clustering, and classification tasks into a uniform learning framework. 
The framework comprises an autoencoder neural network structure to extract features from the raw signals and form a compressed latent feature representation. Furthermore, it contains a k-Means clustering algorithm to partition the unlabeled dataset into groups to produce pseudo labels for the instances. Finally, it contains a DNN classifier to train the human activity classification model based on the latent features and pseudo labels. 
The authors conducted extensive experiments on three publicly available datasets showing that the proposed approach outperforms existing clustering methods under completely unsupervised conditions and achieves a performance similar to fully supervised learning when retraining the extracted latent feature representation. 

In \cite{qi2022dcnn}, Qi et al. proposed a framework for smartphone-based HAR that combines data from the Microsoft Kinect camera and the smartphone's IMU signals to identify 12 complex daily activities. The proposed framework comprises five clustering layers and a DL-based classification model. The authors employed a hierarchical k-medoids (Hk-medoids) algorithm to obtain labels with a high accuracy. 
Additionally, the performance of a deep convolutional neural network (DCNN) classification model was evaluated and compared to other ML and DL methods. Moreover, the authors proposed a calibration approach to mitigate the effect of artifact and drifting noise on the obtained 3D skeleton joints data.

Finally, Lin et al.~\cite{lin2022clustering} designed a feature selection technique based on Fuzzy C-means particle swarm optimization (FS-FCPSO) to annotate six human activities automatically. 
The results of this method were compared with those of k-means and fuzzy C-means algorithms. The authors used a dataset that included 30 volunteers aged 19 to 48 and captured 3-axis linear acceleration and 3-axis angular velocity using a Samsung Galaxy S II smartphone with an embedded accelerometer and gyroscope. 
The FS-FCPSO method was more suitable for automatic labeling in HAR than the k-means and fuzzy C-means algorithms. 
The main contribution of this research was to adopt a feature selection method based on fuzzy C-average particle swarm optimization (PSE) to improve the accuracy of automatic labeling results. 
The authors reduced 561 feature to 163 features as a cluster subset and showed that feature selection based on binary PSO could effectively enhance the application of the fuzzy C-means clustering method in automatic labeling for HAR.

\subsubsection{Environment-driven approaches}\label{sec:4.2.2}
This section concludes the description of the categories introduced by the taxonomy given in Figure~\ref{fig:tax_2} by discussing methodologies presenting fully automated, environment-driven techniques. Tables~\ref{tab:H-ED} and~\ref{tab:H-ED-1} offer a comprehensive summary of the 5 articles included in this category.

\begin{table*}[!th]
\centering
\caption{Fully-automated environment-driven approaches.}\label{tab:H-ED}
\resizebox{\textwidth}{!}{
\begin{tabular}{c||c
ccccccccc}
\hline
\makecell{\textbf{Ref.}\\ \textbf{(Year)}} & \makecell{\textbf{Dataset}\\\textbf{Provided (\#)}} & \textbf{Devices} & \textbf{Sensors} & \makecell{\textbf{\# of}\\ \textbf{Sensors}} & \textbf{Position} & \textbf{Model} & \makecell{\textbf{\# of} \\\textbf{Subjects [M/F]}} & \textbf{Activities} & \makecell{\textbf{\# of} \\\textbf{Activities}} & \makecell{\textbf{In-Out}\\ \textbf{Door}}\\ \hline \hline

\makecell{\cite{szewcyzk2009annotating}\footref{fo:54}\\ (2009)}
&yes (1)
&environment
&\makecell{motion sensors, \\temperature}
&27
&\makecell{phone book, cooking pot, \\medicine container, \\cooking ingredients pot}
&\textit{n/a}
&2 [2/0]
&OBI-ADL
&6
&indoor\\ \hline

\makecell{\cite{loseto2013mining}\\ (2013)}
&yes (1)
&\makecell{smartphone,\\ mobile app}
&\makecell{accelerometer\\ GPS, mobile app}
&3
&\textit{n/a}
&\makecell{semantic web\\ language}
&1
&\makecell{PA}
&5
&\makecell{indoor\\ outdoor}\\ \hline

\makecell{\cite{al2017annotation}\\ (2017)}
&no (3)
&enviromental
&\makecell{accelerometer,\\ temperature,\\ PIR}
&\makecell{20 - 25\\ - 33}
&\textit{n/a}
&\makecell{HMM,\\ CRF} 
&4 - 2 - 1
&\makecell{ADL}
&16  - 10 - 15
&indoor\\ \hline

\makecell{\cite{demrozi2021towards}\\ (2021)}
&yes (1)
&\makecell{smartwatch,\\ BLE beacons}
&\makecell{accelerometer,\\ gyroscope,\\ magnetometer,\\ BLE antenna}
&2
&wrist, ODLs
&\makecell{regression\\ models} 
&1
&\textit{n/a}
&\textit{n/a}
&indoor\\ \hline

\makecell{\cite{dissanayake2021indolabel}\\ (2021)}
&yes (1)
&\makecell{smartwatch,\\ access points}
&\makecell{accelerometer,\\ gyroscope, RSSI}
&36
&\makecell{wrist,\\ environment AP}
&\makecell{location specificity\\ measure and activity\\ similarity matrix\\ calculation}
&4 [3/1]
&\makecell{ADL, OBI}
&14
&Indoor\\\hline \hline
\multicolumn{11}{c}{In Columns Devices, \# of Sensors, Position, \# of Subjects [M/F], Activities, and \# of Activities the symbol \"-\" is used to separate information of the different datasets.}
\end{tabular}}
\end{table*}

\begin{table*}[!th]
\centering
\caption{Annotation method, Advantages, and Limitations of Fully-automated environment-driven approaches.}\label{tab:H-ED-1}
\resizebox{\textwidth}{!}{
\begin{tabular}{c||
p{7cm}
p{5cm}
p{5cm}}
\hline
\textbf{Ref. (Year)} & \textbf{Annotation} & \textbf{Advantages} & \textbf{Limitations} \\ \hline \hline
\makecell{\cite{szewcyzk2009annotating}\footref{fo:54}\\ (2009)}
& Four distinct annotation systems were employed, each with its own approach: raw data and hard-coded recognition, raw data and resident time diaries, visualization of the sensor data tool, and visualization of the sensor data tool combined with residents' feedback.
& Explored the topic and conducted a preliminary comparison of various annotation systems.
& Recogniton not performed and requires human labels and intervention\\ \hline

\makecell{\cite{loseto2013mining}\\ (2013)}
&Semantic web languages to perform automated profile annotation based on the data collected by embedded micro-devices, logs, and applications on a smartphone.
&Achieves an annotation accuracy of 98\%. 
&Less accurate in indoor environments. \\ \hline

\makecell{\cite{al2017annotation}\\ (2017)}
& Probabilistic modeling of activities through the combination of spatial and temporal relationships, and the algorithmic segmentation of incoming actions.
& No participant intervention is required. Able to extract hidden activities/behaviors.
& The average annotation accuracy is 73\%. \\ \hline

\makecell{\cite{demrozi2021towards}\\ (2021)}
&Recognize the distance between ODLs and a smartwatch through evaluating BLE signals. 
&Fully automated
&BLE signals are strongly influenced by the environment. Method not tested for automatic annotation.
\\ \hline

\makecell{\cite{dissanayake2021indolabel}\\ (2021)}
& Predicts the room location of a user by discovering location-specific sensor data motifs from the user's smartwatch.
& Achieves an F1-Score of 85\% in labeling activities.
& - \\ \hline \hline
\end{tabular}}
\end{table*}

The earliest approach in this a category was proposed by Loseto et al. in~\cite{loseto2013mining}. 
The authors proposed an agent running on an Android mobile app that utilizes semantic web languages to perform automated profile annotation based on the data collected by embedded micro-devices, logs, and applications on a smartphone. 
The system annotates the data by using motion, location, and smartphone usage to annotate the user's activity automatically. The resulting semantic-based daily profile can be leveraged in an ambient intelligence scenario to adapt the environment to user preferences. 

In~\cite{al2017annotation}, Al Zamil et al. proposed a methodology for automated data annotation in smart home environments, specifically for modeling activities based on spatially recognized actions and validating the assignment of labels through temporal relations. 
The proposed technique utilized Hidden Markov Models (HMM) and Conditional Random Field (CRF) models to accurately detect segment labels. 
The authors defined the segmentation problem as an optimization problem that minimizes the ambiguity to improve the overall accuracy. The experiments that were performed on the CASAS data sets~\cite{cook2009collecting} indicated that the proposed methodology achieved a better performance than state-of-the-art methodologies, with contributions including the modeling of activity actions as states and transitions, the incorporation of spatial and temporal relationships, and the algorithmic segmentation of incoming actions. 

In the approach presented by Demrozi et al.~\cite{demrozi2021towards}, BLE beacons were mapped to the locations or objects where a human subject typically performs activities, such as cooking or working. Furthermore, the data collected by sensors embedded in the user's smartwatch were associated to the nearest BLE beacon. This allows data from the smartwatch sensors to be automatically labeled with the human activity that corresponds to the closest beacon. 
The proposed methodology is low-cost and uses regression models to estimate the distance between the user and the beacons accurately. 
The methodology was found to estimate the distance between emitters and receivers with an RMSE of 13 cm and an MAE of 10 cm. The outcome is an automatically-annotated dataset that can be used to design dedicated HAR models. 

Finally, in \cite{dissanayake2021indolabel}, Dissanayake et al. present IndoLabel, a method to automatically detect short sensor data motifs specific to a location class, and builds an environment-independent location classifier without requiring handcrafted rules and templates. 
The authors state that this method can be utilized to extract class-specific sensor data segments from any type of time series sensor data, and can assign semantic labels to any WiFi cluster in daily life, e.g., in hospitals and factories. The authors evaluated the proposed method in real house environments using a leave-one-environment-out cross-validation methods, and achieved state-of-the-art performance despite the unavailability of labeled training data in the target environment.

\subsection{Classification based on employed sensing device}\label{sec:4.3}
As can be inferred from the previous sections, sensing devices play a pivotal role in HAR and related data annotation techniques. We hence in this section classify the reviewed approaches based on their sensing devices used.

Sensing devices, ranging from wearable sensors to environmental sensors, enable the collection of essential data that provide insights into individuals' activities and behavior patterns. By accurately capturing information such as motion, location, heart rate, and environmental context, sensing devices serve as the foundation for HAR systems. The data collected by these sensing devices serve as the raw material for data annotation techniques in HAR. 
Sensing devices facilitate data annotation techniques in several ways. Firstly, they provide objective and quantitative measurements of various physical and environmental parameters, ensuring the accuracy and reliability of the annotated data. This reliability is essential for developing robust HAR models.
Secondly, sensing devices offer real-time or near real-time data, allowing for immediate feedback and annotation during data collection. This feature is particularly valuable in scenarios where prompt intervention or feedback is necessary, such as monitoring athletic performance or tracking a patient's rehabilitation progress. 
Furthermore, sensing devices allow for the annotation of contextual information, such as the location and environmental conditions during specific activities. This additional contextual data enriches the understanding of human behavior and contributes to more nuanced and comprehensive activity recognition models.

To underscore the significance of sensing devices, Table \ref{tab:category} offers a comprehensive overview of the various sensing devices and associated sensors employed in the methodologies and data annotation techniques explored in the preceding sections. 
\begin{table*}[!ht]
\centering
\caption{Categorization of Data Annotation Techniques based on Automation and Data Collection sensor.}\label{tab:category}
\resizebox{0.975\textwidth}{!}{
\begin{tabular}{c|c||c|c|c||c|c|c|}
\cline{3-8}
\multicolumn{2}{c}{}&\multicolumn{6}{|c|}{\textbf{Data annotation techniques in HAR}}\\
\cline{3-8}
\multicolumn{2}{c}{} &
\multicolumn{3}{|c||}{\textbf{Semi-automated}} &
\multicolumn{3}{c|}{\textbf{Fully-automated}} \\ \hline 
\textbf{Sensing Device} & 
\textbf{Involved Sensors} & 
\textbf{Data-driven} & \textbf{Environment-driven} & \textbf{Hybrid}& 
\textbf{Data-driven} & \textbf{Environment-driven} & \textbf{Hybrid}\\
\hline \hline
Video         & Images &
\cite{martindale2018smart} \cite{kwon2020imutube} \cite{avsar2021benchmarking} \cite{korpela2021reducing} &
&
\cite{alam2015mobeacon} \cite{nino2016scalable} & 
\cite{jardim2016automatic} \cite{zhang2020deep} \cite{qi2022dcnn} 
&& \\ \hline
Inertial & \makecell{Accelerometer\\Gyroscope\\ Magnetometer} &

\makecell{
\cite{saeedi2017co} \cite{martindale2018smart} \cite{gan2018automatic} \cite{bota2019semi} \\ 
\cite{martindale2019hidden} \cite{ponnada2019designing} \cite{faridee2019augtoact} \cite{hossain2019active} \\ \cite{sawano2020annotation} \cite{kwon2020imutube} 
\cite{tang2021selfhar} \cite{yhdego2022fall} \cite{mohamed2022har}} &

\cite{subramanya2012recognizing} \cite{woznowski2017talk} \cite{tonkin2018talk} \cite{solis2019human}&

\cite{alam2015mobeacon} \cite{meurisch2015labels}  \cite{cruciani2018automatic} \cite{cruz2019semi} & 

\makecell{\cite{rokni2018autonomous} \cite{liang2018automatic} \cite{zhang2020deep} 
\cite{jeng2021wrist} \\ \cite{ullrich2021detection} \cite{ma2021unsupervised} \cite{qi2022dcnn} \cite{lin2022clustering}} & 

\cite{loseto2013mining} \cite{al2017annotation} \cite{demrozi2021towards} \cite{dissanayake2021indolabel} & 
\\ \hline

Environmental &  \makecell{Pressure\\ Temperature\\ Brightness\\ GPS\\ Microphone\\ PIR} & 
\cite{martindale2018smart} \cite{martindale2019hidden} &
\cite{subramanya2012recognizing} \cite{woznowski2017talk} \cite{tonkin2018talk} & 
\cite{cruciani2018personalized} \cite{meurisch2015labels}  \cite{cruz2019semi}  \cite{tonkin2019towards}& 
& 
\cite{loseto2013mining}\cite{al2017annotation}&
\\ \hline

Physiological &  \makecell{Heart rate\\ Oxygen\\ Blood pressure} &
&
\cite{mohamed2022har} & 
&
& 
&
\\ \hline

Radio signals &  \makecell{Bluetooth\\ WiFi}& &
&
\cite{alam2015mobeacon} \cite{meurisch2015labels}&
&
\cite{demrozi2021towards} \cite{dissanayake2021indolabel}&
\\ 
\hline
\hline
\end{tabular}}
\end{table*}
By providing a structured representation of the devices and sensors used, the table highlights their crucial role in capturing and annotating data in the defined annotation categories. 
Empirical evidence demonstrates that among the different sensing devices employed, inertial sensors have emerged as the most widely utilized for data collection and annotation purposes. Inertial sensors, which encompass accelerometers, gyroscopes, and magnetometers, offer the ability to measure and record an individual's motion, orientation, and spatial positioning. Their popularity can be attributed to their versatility, portability, and ability to provide real-time and fine-grained data. 
The inherent advantages of inertial sensors have positioned them as a dominant choice for researchers and practitioners in the field, facilitating accurate and reliable data annotation for a wide range of applications.

%% file: sec/5-disc.tex
\section{Discussion}\label{sec:disc}
\noindent \textbf{Data Annotation Significance in HAR:} In recent years, as shown by our searching strategy, there has been a notable surge in studies investigating methods for data annotation in HAR. This trend reflects the growing recognition of the importance of accurate and efficient annotation techniques in extracting meaningful insights from individuals' daily life activities.
Such rising interest is driven by the recognition of its potential applications in various domains, such as healthcare, smart environments, and personalized services. 
Nevertheless, the complexity of daily life activities and the challenge of collecting and annotating corresponding data are intertwined in a mutually reinforcing manner. As individuals go about their routines, the range and intricacy of activities they engage in can be overwhelming. From personal tasks like commuting, shopping, and exercising to professional responsibilities, social interactions, and leisure pursuits, the spectrum of daily activities is vast. Each activity comprises numerous elements, such as time, location, duration, and context, which need to be captured accurately to gain a comprehensive understanding of an individual's life. 
Thus, collecting and annotating such data poses significant complexities. 
In addition, the diversity of data sources, including smartphones, wearables, and environmental sensors, and the subjective nature of annotating data, such as categorizing activities and determining their significance, introduces inherent biases and uncertainties. \\
\begin{table*}[!th]
\centering
\caption{A comprehensive examination of the features associated with Data Annotation Techniques for Human Activity Recognition (HAR).}\label{tab:annot_2}
\resizebox{0.9\textwidth}{!}{
\begin{tabular}{c||p{8cm}|p{8cm}}
\cline{2-3}
&
\textbf{Fully Automated}  & 
\textbf{Semi Automated}\\ \hline\hline
\textbf{Advantages} & 
$\bullet$ Scalability and consistency \newline
$\bullet$ Speeds up the annotation process \newline
$\bullet$ Recognizes complex patterns &
$\bullet$ Higher precision and adaptability \newline
$\bullet$ Deeper comprehension of the data \newline
$\bullet$ Captures a variety of activities more accurately \newline
$\bullet$ Human-in-the-loop approaches can improve accuracy and ensure quality control \\\hline
\textbf{Disadvantages}&
$\bullet$ Potential lower accuracy compared to semi-automated methods \newline
$\bullet$ Trade-offs between speed and accuracy \newline
$\bullet$ Data quality challenges, as fully automated methods are more susceptible to biases and errors in the training data \newline
$\bullet$ Interpretability and transparency challenges, as the underlying decision-making processes of the algorithm may not be easily understood by humans \newline
$\bullet$ Higher resource requirements, such as computing power and specialized software or hardware &
$\bullet$ Requires involvement of human experts \newline
$\bullet$ Time-consuming and resource-intensive \newline
$\bullet$ Data quality challenges, as the accuracy of semi-automated methods is dependent on the quality of human annotations \newline
$\bullet$ Interpretability and transparency challenges, as the algorithms used in semi-automated methods may not be easily understood by non-experts \newline
$\bullet$ Resource requirements depend on the level of human involvement in the annotation process
\\\hline
\makecell{\textbf{Performandce} \\\textbf{trade-offs}} &
$\bullet$ Potential lower accuracy compared to semi-automated methods \newline
$\bullet$ Trade-offs between speed and accuracy &
$\bullet$ Requires involvement of human experts \newline
$\bullet$ Time-consuming and resource-intensive \\\hline
\makecell{\textbf{Human-in-the-} \\\textbf{loop approaches}} &
$\bullet$ Limited involvement of human experts &
$\bullet$ Active involvement of human experts \newline
$\bullet$ Can improve accuracy and ensure quality control \\\hline
\makecell{\textbf{Data quality} \\\textbf{challenges}}&
$\bullet$ More susceptible to biases and errors in the training data \newline
$\bullet$ Data quality challenges are amplified by the scale and complexity of fully automated annotation methods &
$\bullet$ Accuracy of semi-automated methods is dependent on the quality of human annotations \newline
$\bullet$ Challenges with data quality can arise if the human annotators are not sufficiently trained or if there is disagreement among annotators \\\hline
\makecell{\textbf{Interpretability} \\\textbf{and transparency}}&
$\bullet$ May not be easily understood by humans \newline
$\bullet$ Challenging to achieve transparency \newline
$\bullet$ Explainability may be difficult to achieve &
$\bullet$ Can be easier to understand by humans \newline
$\bullet$ Offers a higher level of transparency and explainability \newline
$\bullet$ Interpretability is enhanced by the involvement of human experts \\\hline
\makecell{\textbf{Resource} \\\textbf{requirements}}&
$\bullet$ Higher computing power and specialized software or hardware may be required  \newline
$\bullet$ More expensive to implement and maintain &
$\bullet$ Resource requirements depend on the level of human involvement in the annotation process \newline
$\bullet$ Can be less expensive to implement and maintain compared to fully automated methods, especially if annotation tasks are simple and require less human input
\\ \hline\hline
\end{tabular}}
\end{table*}

\noindent \textbf{Our perspective:} To this end, deciding between fully automated and semi-automated data annotation techniques is crucial in HAR. Fully automated methods offer scalability and the ability to process large volumes of data, but they may lack accuracy and transparency. In contrast, semi-automated methods combine machine learning models with human expertise, providing higher precision, adaptability, and a deeper understanding of the data. 

Fully automated techniques excel in speeding up the annotation process and detecting complex patterns, but they may be less accurate and suffer from interpretability issues. They heavily rely on software (algorithms or machine learning models) or hardware support (sensors or smart devices), making it challenging to explain their annotations. 
However, such techniques are more prone to biases and errors in the training data and require significant computing power and specialized software or hardware. 
Instead, semi-automated methods leverage machine learning models and human experts, resulting in higher precision, adaptability, and a better understanding of the data. Involving human annotators improves the accuracy and quality, but, demands time and resources due to their involvement, and the accuracy depends on their capabilities. Despite these challenges, semi-automated methods offer interpretability and transparency because human experts contribute to decision-making.

To choose the appropriate automated data annotation approach for human activity recognition, it is essential to consider the advantages and disadvantages of fully automated and semi-automated methods. Factors such as performance trade-offs, human-in-the-loop approaches, data quality challenges, interpretability and transparency, and resource requirements play a significant role in selecting the most suitable method based on specific demands, goals, available resources, and existing knowledge.

In this paper, we categorized these approaches into data-driven, environment-driven, and hybrid methods, based on their underlying principles and methodologies. 
Data-driven techniques rely on data characteristics for annotation, while environment-driven techniques consider the context and environment in which the data was collected. 
Hybrid methods combine both approaches, aiming for more accurate and robust results by integrating data-driven analysis with contextual information. 
Categorizing methods into data-driven, environment-driven, and hybrid approaches allows for informed decision-making, helping researchers and practitioners to select the most suitable approach that aligns with their objectives and requirements.\\

\noindent \textbf{Advantages and Disadvantages:} To summarize, Table \ref{tab:annot_2} provides an overview of the characteristics that need to be considered when designing a new methodology for data annotation in HAR.\\


\noindent \textbf{Recent trends:}  Moreover, new techniques (i.e., Zero-shot, few-shot, and self-supervised learning) for data annotation are finding their space. In particular, Zero-shot learning techniques address the problem of annotating instances or activities that are not included in the training data. This involves identifying and categorizing activities that were only partially observed or not observed at all during the training phase. By leveraging prior knowledge and auxiliary information about related activities, zero-shot learning enables the annotation system to generalize and make accurate predictions for unobserved classes or activities \cite{al2020zero}.

In addition, few-shot learning enables the annotation system to learn from a small number of annotated instances instead of requiring a large amount of annotated data for each activity. This is especially useful when obtaining a large annotated dataset for all possible activities is difficult or time-consuming. Few-shot learning allows the annotation system to generalize from limited labeled data to annotate new instances or activities accurately \cite{tseng2022haa4d}.

Furthermore, self-supervised learning algorithms in HAR utilize unlabeled data's intrinsic structure or information to discover meaningful representations. 
These learned representations can then be utilized to enhance the precision and efficiency of activity recognition during the annotation process. Self-supervised learning enables the annotation system to maximize available data and employ innate knowledge to drive the annotation process \cite{saeed2019multi}.
Consequently, incorporating zero-shot learning, few-shot learning, and self-supervised learning techniques into HAR annotation systems makes it possible to annotate a broader range of activities, handle limited annotated data and increase the system's adaptability to diverse scenarios and activity recognition tasks, thereby expanding its capabilities.

%% file: sec/6-conc.tex
\section{Conclusion}\label{sec:conc}
\begin{figure*}[!htb]
\centering
\includegraphics[width=0.8\textwidth,page={5}]{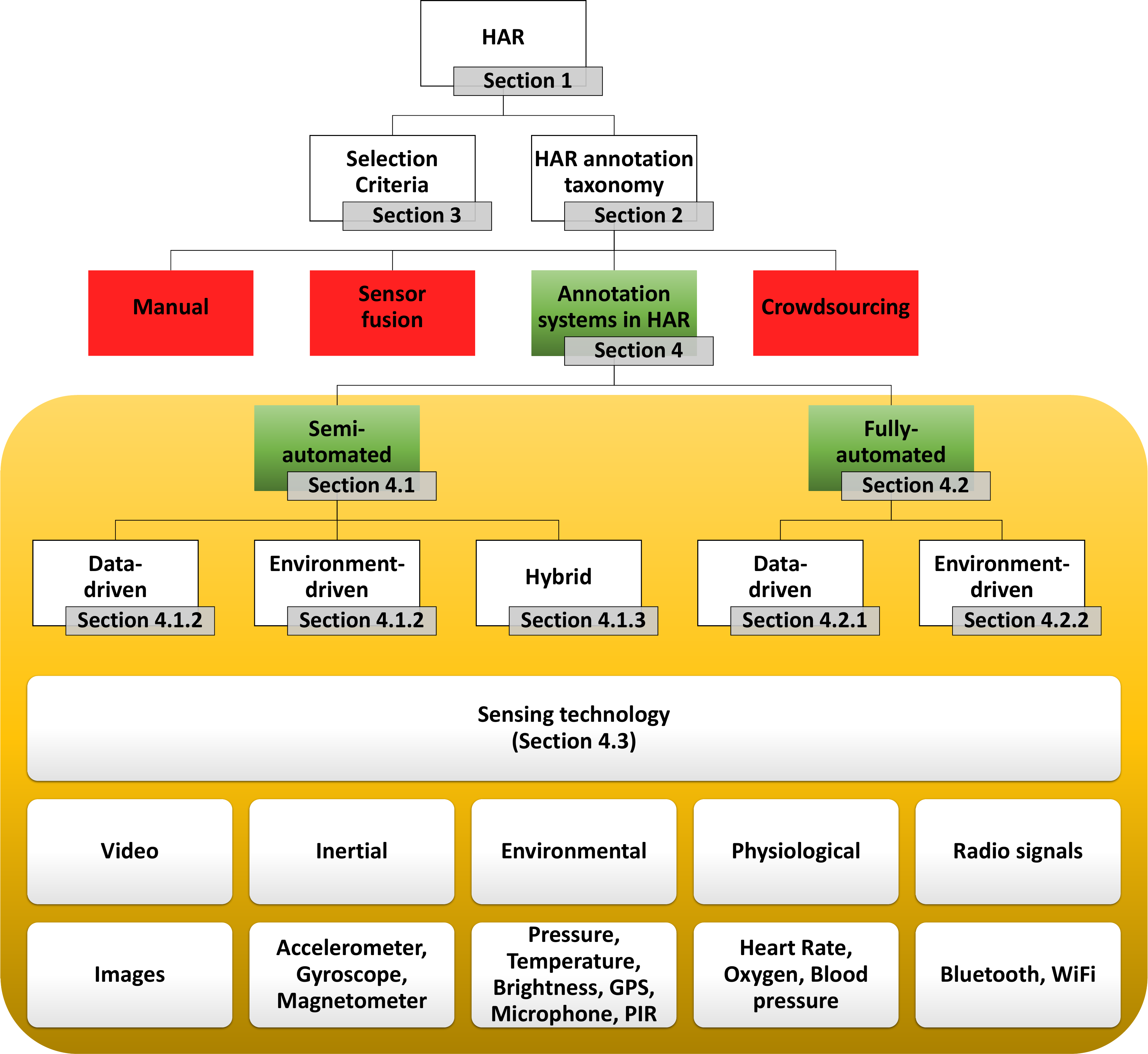}
\caption{Overview of the proposed survey structure on (Semi) Automatic data annotation techniques in HAR from 01/01/1980 to 21/01/2023.}\label{fig:overall}
\end{figure*}

In conclusion, the complexity of daily life activities and the intricacies of collecting and annotating relevant data create a multifaceted challenge in HAR (Section \ref{sec:intro}). As summarized in Figure \ref{fig:overall}, this paper presents the first systematic review about (Semi-) Automatic data annotation techniques in HAR from 01/01/1980 to 21/01/2023 (Section \ref{sec:search}). 
In particular, concerning the HAR annotation taxonomy introduced in Section~\ref{sec:back}, different approaches have been exploited, e.g., manual, sensor fusion, semi-automated (Section \ref{sec:4.1}), fully-automated (Section \ref{sec:4.2}), and crowdsourcing. 
The decision between fully automated and semi-automated data annotation techniques is crucial in addressing this challenge.

Fully automated methods (Section \ref{sec:4.2}) provide scalability and the ability to quickly process large volumes of data. 
They excel in detecting complex patterns but may be less accurate and suffer from interpretability issues.
On the other hand, semi-automated methods (Section \ref{sec:4.1}) leverage machine learning models and human expertise, resulting in higher precision, adaptability, and a better understanding of the data. Involving human annotators thereby improves accuracy and quality control. 

Choosing the appropriate automated data annotation approach for HAR requires considering factors such as performance trade-offs, human-in-the-loop approaches, data quality challenges, interpretability and transparency, and resource requirements. Both fully automated and semi-automated methods can be developed in a data-driven  (Section~\ref{sec:4.1.1} and Section~\ref{sec:4.2.1}), environment-driven (Section~\ref{sec:4.1.2} and Section~\ref{sec:4.2.2}), or hybrid (Section \ref{sec:4.1.3}) manner. The decision depends on the application's demands, goals, available resources, and existing knowledge. Besides, when exploiting the annotation system, the decision must also consider the used sensing technology (Section~\ref{sec:4.3}). 

All approaches have shown promising results in reducing the amount of work and time to be spent on data annotation, while maintaining the annotation accuracy. However, the choice between fully automated and semi-automated methods should be based on specific demands, goals, available resources, and knowledge. Understanding the advantages and limitations of each approach enables informed decision-making, allowing researchers and practitioners to select the most suitable method that aligns with their objectives and requirements.

Finally, the use of zero/few-shot learning and self-supervised learning as part of HAR can potentially improve the practicality and application of annotation systems in real life. HAR systems may be implemented in more contexts and fields, if the scope of actions is limited. This paves the way for using HAR in industries like healthcare, sports analytics, intelligent settings, and surveillance.